%% file: main_arxiv.tex
\documentclass{fairmeta}

\input{math_commands.tex}

\usepackage{nicefrac}
\usepackage{pifont}
\usepackage{xspace}
\usepackage{xurl}
\usepackage{enumitem}
\usepackage{array}
\usepackage{makecell}
\usepackage{longtable}
\usepackage{wrapfig}
\newcommand{\needlines}[1]{\par\ifdim\dimexpr\pagegoal-\pagetotal\relax<#1\baselineskip\newpage\fi}
\BeforeBeginEnvironment{wraptable}{\needlines{22}}
\let\arxivWrapFigure\wrapfigure
\let\arxivEndWrapFigure\endwrapfigure
\RenewDocumentEnvironment{wrapfigure}{O{} m O{} m}
  {% Figure 6 follows an existing 16-line guard before its heading.
   \ifnum\value{figure}=5\needlines{14}\else\needlines{18}\fi
   \arxivWrapFigure{r}{#4}}
  {\arxivEndWrapFigure}

\newcommand{\sys}{ServeLearnBench\xspace}

\newcolumntype{P}[1]{>{\raggedright\arraybackslash}p{#1}}

\renewcommand\affiliation[2][]{\addtolist[#1]{#2}{\affiliationlist}{\ttfamily\affiliationformat}{\quad\textperiodcentered\quad}}
\renewcommand\authorformat[2][]{{\ttfamily #2$^{#1}$}}

\title{{\fontsize{17.28pt}{22pt}\selectfont
\sys: How Well Can Agents Self-Improve from Serving Experience?}}
\author[1]{Haizhong Zheng}
\author[1]{Yizhuo Di}
\author[1]{Ranajoy Sadhukhan}
\author[2]{Shuowei Jin}
\author[1]{Beidi Chen}
\affiliation[1]{Carnegie Mellon University}
\affiliation[2]{Amazon}

\metadata[GitHub]{\url{https://github.com/Infini-AI-Lab/ServeLearnBench}}
\metadata[Website]{\url{https://infini-ai-lab.github.io/ServeLearnBench}}

\hypersetup{
  pdftitle={ServeLearnBench: How Well Can Agents Self-Improve from Serving Experience?},
  pdfauthor={Haizhong Zheng, Yizhuo Di, Ranajoy Sadhukhan, Shuowei Jin, Beidi Chen}
}

\abstract{
Large language model agents are increasingly deployed to perform complex tasks in real-world environments. However, the knowledge required for correct behavior in these environments is often implicit, undisclosed, and subject to change over time. Recent continual-learning harnesses seek to address this challenge by enabling agents to improve from serving experience. Yet the effectiveness and limitations of these methods are not yet well characterized. Existing benchmarks provide only partial coverage: some explicitly provide the target knowledge, others assume a static environment, and those that support continual adaptation remain limited in scale and knowledge diversity. To enable systematic evaluation, we formalize an \emph{evolving-environment streaming dataset} (EESD), in which agents must infer, apply, and revise latent environment knowledge from interaction and outcome feedback as hidden policies evolve, and introduce \sys, spanning retail support, banking, and sales-pitch generation with 53 environment windows and 7,718 tasks. We evaluate five learning harnesses (RAG, Mem0, SkillOpt, Continual Harness, and Prime) across six models (GPT-5.6 Terra, Opus 5, Kimi K3, GLM-5.3, DeepSeek V4.1 Flash, and GLM-5.3 Flash), covering 28 model--harness pairs and 252 learning runs. Our evaluation reveals three main findings: a substantial gap remains between task capability and learning from experience; continual adaptation is costly and can degrade already-correct behavior; and insufficient exploration emerges as a key bottleneck to effective adaptation. Overall, \sys provides a controlled testbed for diagnosing these limitations and tracking progress toward agents that continually and reliably improve through serving experience.
}

\begin{document}
\maketitle

\begin{figure*}[h!]
  \centering
  \includegraphics[width=\textwidth]{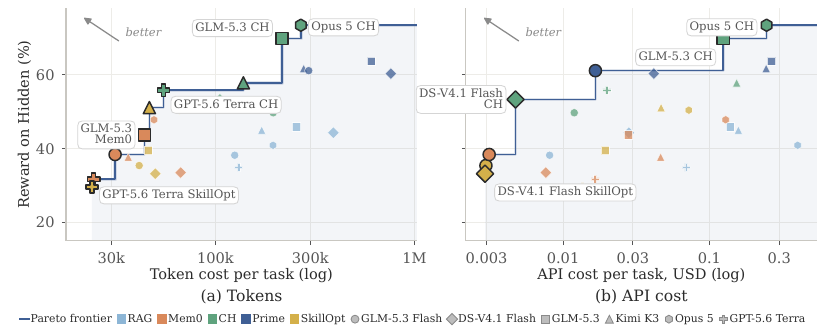}
\caption{\textbf{Cost vs.\ Performance.}
\textbf{(a)} Hidden-case reward versus token cost per test task (prefill, KV-cache hits and output tokens included).
\textbf{(b)} Hidden-case reward versus API cost per test task.
Reward is the Avg.~H of Table~\ref{tab:main-results} (mean over the nine settings).
Lines connect Pareto-efficient pairs, for which no other plotted pair achieves both lower cost and higher reward.}
\label{fig:cost-frontier}
\end{figure*}

\clearpage
\input{tex/01_introduction}
\input{tex/02_related_work}
\input{tex/03_benchmark}
\input{tex/04_experimental_setup}
\input{tex/05_results}
\input{tex/07_conclusion}

% \clearpage
% \input{tex/08_statements}
\bibliographystyle{plainnat}
\bibliography{references}

\clearpage
\beginappendix
\input{tex/a_overview}
\input{tex/a_additional_experimental_setting}
\input{tex/a_benchmark_details}

\input{tex/a_additional_evaluation}
\end{document}

%% file: math_commands.tex
\usepackage{amsmath,amsfonts,bm}

\def\eqref#1{equation~\ref{#1}}
\def\1{\bm{1}}

\DeclareMathAlphabet{\mathsfit}{\encodingdefault}{\sfdefault}{m}{sl}
\SetMathAlphabet{\mathsfit}{bold}{\encodingdefault}{\sfdefault}{bx}{n}

%% file: tex/01_introduction.tex
\section{Introduction}
\label{sec:intro}
\vspace{-0.1cm}

% Alternative opening: retain the original paragraphs below for comparison.
Language agents increasingly handle complex real-world workflows~\citep{yao2024taubench,zhou2024webarena,wei2025swerl,jimenez2023swe,zheng2025deepresearcher,gao2025beyond}, but deployment is not a one-time test of capability. Operational policies, system constraints, and user preferences may change over time, often without corresponding updates to the information available to the agent. Reliable service therefore requires agents to learn from interaction and failure rather than continue acting on an increasingly outdated model of their environment.
Recent continual-learning harnesses, including SkillOpt~\citep{yang2026skillopt}, Continual Harness~\citep{karten2026continualharness}, and Prime~\citep{karten2026primeagent}, aim to support this process by updating persistent memories, skills, or strategies from experience. 
Yet the ability to update an agent does not necessarily imply that it has acquired useful, reusable knowledge. 
This raises a central question: \emph{how effectively can these systems infer previously unknown environment knowledge from serving experience, revise it as conditions change, and maintain reliable behavior over time?}

Understanding this capability requires answering three questions: 
\textbf{1)} How well do current agents learn continually from serving experience? \textbf{2)} What does this learning cost, in both computation and existing capabilities? 
\textbf{3)} What enables effective continual learning? Existing benchmarks have begun to study parts of this problem~\citep{cheng2023llfbench,wu2024streambench,zheng2025lifelongagentbench,shu2026agentcl,guan2026continualskillbench,asawa2026clbench}, but leave important gaps. 
Some \emph{explicitly provide the target knowledge}, testing whether agents can retain and apply it rather than discover it through interaction and failure~\citep{maharana2024locomo,wu2025longmemeval,hsieh2024ruler}. 
Others require agents to discover hidden knowledge but assume \emph{static environments}~\citep{he2025idea,jansen2024discoveryworld,zheng2026herosjourney}, while benchmarks that support changing environments remain limited in \emph{scale and knowledge diversity}~\citep{asawa2026clbench}. These limitations make it difficult to systematically measure how agents acquire and revise hidden knowledge over time, quantify the costs of adaptation, and identify the mechanisms that make continual learning effective.

% In real-world deployment, agents must continually learn from serving experience while adapting to environments whose rules and conditions are only partially known and may change over time.
An ideal benchmark for continual learning from serving experience should satisfy three criteria.
First, it should require \emph{learning from interaction and failure}: agents must infer the knowledge needed for future tasks from their previous interactions with the environment, including failed attempts and the resulting feedback, without being explicitly told that knowledge.
Second, it should feature \emph{evolving environments}: policies, constraints, or preferences should change during the task stream, requiring agents to recognize when learned knowledge becomes outdated and adapt their behavior using new experience.
Third, it should offer \emph{sufficient scale}: many tasks involving different kinds of environment knowledge should support systematic analysis of how agents acquire, retain, apply, and revise that knowledge over extended serving histories.

\begin{figure}[t]
  \centering
  \includegraphics[width=0.90\linewidth]{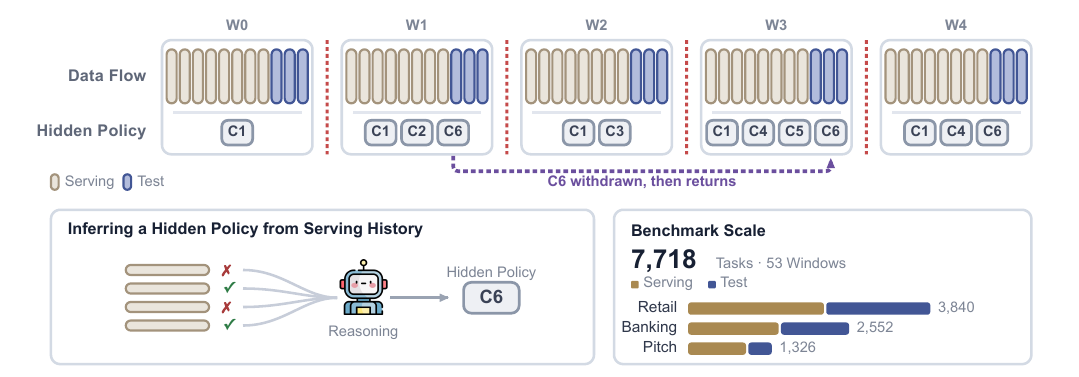}
\caption{\textbf{ServeLearnBench protocol.}
\emph{Top:} the stream is divided into environment windows, each containing feedback-provided serving tasks followed by feedback-hidden test tasks. The active policy in window $s$ consists of behavioral components $\mathcal{C}(\theta_s)$, which may be introduced, withdrawn, or reactivated (dashed lines).
\emph{Bottom left:} the learner receives neither policy-component descriptions nor corrections, and must infer the active policy from serving interactions and outcome feedback. \emph{Bottom right:} benchmark scale across domains, tiers, windows, and tasks.}
\vspace{-0.35cm}
  \label{fig:overview}
\end{figure}

In this paper, we formalize this setting as an \emph{evolving-environment streaming dataset} (EESD): a chronological stream of tasks governed by latent environment knowledge that evolves over time.
We introduce \sys as a controlled suite of EESDs (Figure~\ref{fig:overview}).
To enforce \emph{learning from interaction and failure}, the target environment knowledge is never explicitly revealed in task inputs or feedback, agents must instead infer it from their accumulated experience interacting with the environment.
To instantiate \emph{evolving environments}, the hidden policy changes over time rather than remaining fixed, requiring agents to continually adapt their knowledge and behavior.
Finally, to provide \emph{sufficient scale}, \sys comprises 53 environment windows, 4,508 serving tasks, and 3,210 held-out test tasks across retail support, banking case review, and sales-pitch generation, covering categorical rules, numeric and compositional thresholds, and latent preferences.

To evaluate how effectively current continual learning methods learn from serving experience and improve future behavior, we compare five learning harnesses across six models.
We conduct the evaluation on five harnesses~(RAG~\citep{lewis2020rag}, Mem0~\citep{chhikara2025mem0}, SkillOpt~\citep{yang2026skillopt}, Continual Harness (CH)~\citep{karten2026continualharness}, Prime~\citep{karten2026primeagent}) and six models (GPT-5.6 Terra~\citep{openai2026gpt56terra}, Opus 5~\citep{anthropic2026opus5}, Kimi K3~\citep{kimiteam2026k3}, GLM-5.3~\citep{zai2026glm53}, DeepSeek V4.1 Flash~\citep{deepseekai2026v41flash}, and GLM-5.3 Flash~\citep{zai2026glm53flash}).
The evaluation yields three key findings:
\begin{enumerate}[label=\textbf{\arabic*.},leftmargin=1.5em]
\item \textbf{Performance gaps remain in continual learning.} (Section~\ref{sec:main-evaluation})
All models can perform the tasks once the policy is disclosed: Oracle achieves 95.4\% accuracy on Hidden cases on average. Yet no agent learns the policy reliably. On Retail L3, the best of the 28 model--harness pairs reaches only 59.6\%, while the median pair achieves 22.0\%.
\item \textbf{No free lunch for continual learning.} (Sections~\ref{sec:results-run-cost} and~\ref{sec:results-interference})
Harness with continual learning support is more expensive and can degrade already-correct behavior.
Inferring a hidden policy costs 4--101$\times$ as much as following a disclosed one: on Kimi K3, Prime spends \$2{,}479 compared with \$102 for Oracle. Adaptation also hurts performance on tasks that require no learning. 
\item \textbf{Effective exploration is critical for continual learning.} (Section~\ref{sec:results-exploration})
A harness's ability to explore during unsuccessful adaptation is strongly associated with its final performance. Harnesses that vary their candidate behaviors more achieve higher overall reward. This relationship highlights effective exploration as an important design target for future continual-learning harnesses.
\end{enumerate}

%% file: tex/02_related_work.tex
\vspace{-0.3cm}
\section{Related Work}
\vspace{-0.2cm}
\label{sec:related}

\paragraph{Memory and continual-learning benchmarks.}
Interactive-agent benchmarks first established outcome-based evaluation of policy following and tool use; $\tau$-bench is especially relevant because it verifies user interaction, tool execution, and final database state under a given policy~\citep{yao2024taubench}.
Feedback-stream benchmarks then made improvement across tasks measurable: LLF-Bench provides rich language feedback, whereas StreamBench includes continuous input--feedback streams and binary outcomes~\citep{cheng2023llfbench,wu2024streambench}.
LifelongAgentBench, AgentCL, ContinualSkillBench, SkillLearnBench, and CL-Bench extend this view to acquisition, reuse, composition, and retention across sequential tasks and persistent agent states~\citep{zheng2025lifelongagentbench,shu2026agentcl,guan2026continualskillbench,zhong2026skilllearnbench,asawa2026clbench}.
StateMemBench more specifically tests whether memory systems track revised facts and distinguish the current state from superseded ones~\citep{fan2026statemembench}.
Dynamic-environment suites add complementary forms of change: EvoArena~\citep{xu2026evoarena} updates interactive environments, while \citet{harrington2026when} separate performance on drifting and stable knowledge.
EnvHarness instead reshapes static environments into adaptive practice environments while retaining their verifiers~\citep{huang2026envharness}.
These benchmark families emphasize different capabilities rather than sharing one limitation.
What remains only partially combined is an evaluator-controlled evolving environment in which the operative state is hidden, feedback reveals outcomes rather than corrections, prior knowledge can become stale or valid again, and learning gains can be separated from collateral effects through matched controls.
ServeLearnBench targets this intersection with evolving-environment streaming datasets, Blind/Oracle references, and Hidden versus Fully Specified evaluation slices.

\paragraph{Continual learning for self-evolving agents.}
Self-evolving agents differ in the state they update and reuse. RAG and Mem0 retain prior evidence; ReasoningBank distills strategies from successes and failures, WorldEvolver revises predictive rules from observed transitions, and Reflexion and ExpeL preserve verbal lessons~\citep{lewis2020rag,chhikara2025mem0,ouyang2025reasoningbank,zhang2026worldevolver,shinn2023reflexion,zhao2024expel}.
GEPA, RHO, RHI, and SkillOpt revise prompts, harness specifications, or skills using trajectory feedback, while ACE curates context and PolicyBank refines structured policy knowledge~\citep{agrawal2025gepa,pan2026rho,lee2026rhi,yang2026skillopt,zhang2026ace,choi2026policybank}.
Continual Harness and Prime update broader harness state during interaction; HCL examines retention during such updates, Adaptive Auto-Harness routes changing task streams, JIT-Agent generates task-specific harnesses, and RSIAgent explores new environments to build reusable knowledge~\citep{karten2026continualharness,karten2026primeagent,kang2026hcl,liu2026adaptiveautoharness,zhang2026jitagent,zhu2026rsiagent}.
AlphaEvolve and EvoX instead evolve code and search strategies, while controlled harness-design experiments isolate the effects of planning, actions, and context management across models~\citep{novikov2025alphaevolve,liu2026evox,fan2026harnessdesign}.
% Model-parameter RL provides a further route: DeepSeek-R1 and DAPO train reasoning with reward-driven rollouts; GRESO selects informative rollouts, M2PO exploits stale trajectories, and AstraFlow and RLBoost improve the dataflow and resource efficiency of training~\citep{deepseekai2025deepseekr1incentivizingreasoningcapability,yu2025dapo,zheng2025act,zheng2025prosperity,zheng2026astraflow,wu2025rlboost}.
ServeLearnBench evaluates how well agents infer hidden knowledge from serving outcomes, revise it after changes, and preserve behavior on Fully Specified cases under a common task stream.

% \begin{table}[h!]
% \centering
% \small
% \setlength{\tabcolsep}{3.2pt}
% \caption{Representative continual-agent settings emphasize complementary capabilities. \cmark: central design; \pmark: partial or variant-dependent; \xmark: absent in the released evaluation. ``Bound'' denotes an environment-informed empirical upper bound paired with the no-learning lower bound.}
% \label{tab:related}
% \resizebox{\linewidth}{!}{%
% \begin{tabular}{lccccc}
% \toprule
% Benchmark & Sequential & Truth changes & Change hidden & Outcome-only & Bound + slices \\
% \midrule
% StreamBench~\citep{wu2024streambench} & \cmark & \xmark & -- & \cmark & \xmark \\
% LifelongAgentBench~\citep{zheng2025lifelongagentbench} & \cmark & \xmark & -- & \pmark & \xmark \\
% AgentCL~\citep{shu2026agentcl} & \cmark & \xmark & -- & \cmark & \pmark \\
% CL-Bench~\citep{asawa2026clbench} & \cmark & \pmark & \pmark & \pmark & \xmark \\
% PolicyBank~\citep{choi2026policybank} & \cmark & \xmark & -- & \xmark & \pmark \\
% EvoArena~\citep{xu2026evoarena} & \cmark & \cmark & \xmark & \pmark & \xmark \\
% When Does CL Require Learning~\citep{harrington2026when} & \cmark & \cmark & \pmark & \pmark & \pmark \\
% \textbf{ServeLearnBench} & \cmark & \cmark & \cmark & \cmark & \cmark \\
% \bottomrule
% \end{tabular}}
% \end{table}
% \FloatBarrier

%% file: tex/03_benchmark.tex
% \vspace{-0.2cm}
\section{ServeLearnBench}
\label{sec:benchmark}
% \vspace{-0.2cm}

In this section, we present the design of ServeLearnBench.
We first formalize the evolving-environment streaming dataset abstraction (Section~\ref{sec:formal}), then describe task creation, reward metrics, and reference baselines across domains and difficulty tiers (Section~\ref{sec:domains}).

\vspace{-0.2cm}
\subsection{Evolving-Environment Streaming Dataset (EESD)}
\label{sec:formal}

To evaluate continual learning from experience, a benchmark should require agents to infer hidden environment knowledge and adapt when that knowledge changes. We capture this with an \emph{evolving-environment streaming dataset} (EESD), a chronological task stream in which the learner-visible interface remains fixed while the evaluator-private environment evolves over time. This formulation tests if agents can acquire, apply, and revise environment knowledge from prior interactions.

% \paragraph{Definition.}
% An \emph{evolving-environment streaming dataset} (EESD) is a chronologically ordered collection of tasks in which the latent environment state governing correct behavior changes over time while the observable task surface remains fixed.
% For a domain $g$, the observable surface includes the reference document $d_g$, task interface, and available actions; the evaluator-private state lies in $\Theta_g$ and may encode rules, constraints, preferences, or other behavioral requirements.
% A task $z_n$ comprises a request, the task state needed to instantiate it, the agent's interaction trajectory, and its outcome.
% Unlike a collection of independent examples, an EESD requires the learner to infer behaviorally relevant environmental changes from prior experience without corrupting behavior that remains valid.

\paragraph{Stream and environment windows.}
For each domain--tier instance, ServeLearnBench defines an EESD as a chronological sequence of environment windows, $\mathcal{D}=W_0 \,\Vert\, W_1 \,\Vert\, \cdots \,\Vert\, W_{S-1}$, where $\Vert$ denotes temporal concatenation. Each window $W_s$ is governed by a latent environment state $\theta_s\in\Theta_g$ that remains fixed within the window and may change at window boundaries. Across windows, the environment may introduce new policies, revise existing ones, or return to previously active policies. Neither the active state nor its changes are disclosed to the learning harness.

% \paragraph{Stream and environment windows.}
% For each domain--tier instance, ServeLearnBench defines one EESD
% \begin{equation}
%   \mathcal{D}=W_0\,\Vert\,W_1\,\Vert\cdots\Vert\,W_{S-1},
%   \label{eq:stream}
% \end{equation}
% where $\Vert$ denotes temporal concatenation and $W_s$ is an environment window.
% Let $\theta_s\in\Theta_g$ denote the latent environment state for $W_s$.
% The state is constant throughout the complete window,
% \begin{equation}
%   \theta(n)=\theta_s \qquad \text{for }z_n\in W_s,
%   \label{eq:environment-process}
% \end{equation}
% and may change only after the window ends.
% The process may introduce a new state, revise an earlier state, or return to one previously active.
% The latent policy $\theta_s$ and the content of its changes are not disclosed to learning harnesses.

\paragraph{Serving and test splits.}
Each environment window contains serving tasks followed by test tasks, $W_s=\mathcal{D}^{\mathrm{serve}}_s\,\Vert\,\mathcal{D}^{\mathrm{test}}_s$.
Serving and test tasks cover the same behavioral components active under $\theta_s$, while using disjoint task instances. Let $\mathcal{C}(\theta_s)$ denote the set of active components. We require
\begin{equation}
  \operatorname{Cov}(\mathcal{D}^{\mathrm{serve}}_s)
  =
  \operatorname{Cov}(\mathcal{D}^{\mathrm{test}}_s)
  =
  \mathcal{C}(\theta_s),
  \qquad
  \mathcal{D}^{\mathrm{serve}}_s
  \cap
  \mathcal{D}^{\mathrm{test}}_s
  =
  \emptyset.
  \label{eq:matched-coverage}
\end{equation}
This ensures that test tasks evaluate transfer to new instances under the same active policy rather than memorization of serving examples.

% Thus no evaluated rule is absent from the available experience, and no serving example is repeated as a test.

\paragraph{Hidden-Dependent and Fully Specified data.}
Orthogonal to the serving/test split, each task is categorized by whether correct behavior depends on the latent environment state. A task is \emph{Hidden-Dependent} if learner-visible information alone is insufficient to determine the correct behavior, and \emph{Fully Specified} otherwise. Hidden-Dependent tasks measure whether agents acquire and apply hidden environment knowledge, whereas Fully Specified tasks measure whether adaptation preserves behavior that is already supported by visible information.

% Appendix~\ref{app:evaluation-protocol} details the evaluator and learner protocol.

\subsection{ServeLearnBench: Task Creation}
\label{sec:domains}

ServeLearnBench defines nine scenarios by crossing three domains~(Retail, Banking, and Pitch) with three difficulty tiers. Each scenario follows the same EESD protocol while fixing its task interface, verifier, hidden policies, and window schedule:

\paragraph{Retail.}
Retail is a multi-turn support environment covering four core request types: canceling or modifying pending orders, and returning or exchanging delivered items. Tasks vary users, orders, products, payment methods, destinations, and single- versus multi-item requests, with harder variants requiring multi-item resolution or fallback selection. Each task operates on an isolated database and is considered correct only if it reaches the intended final state or produces the correct refusal.

\begin{wraptable}{r}{0.34\textwidth}
\centering
\footnotesize
\setlength{\tabcolsep}{3pt}
\renewcommand{\arraystretch}{0.92}
\caption{
Detailed benchmark statistics (W: number of windows). Appendix~\ref{app:benchmark-details} provides more details.
}
\label{tab:tiers}
\begin{tabular}{@{}lrrr@{}}
\toprule
Scenario & W & Serving & Test \\
\midrule
Retail L1 & 6 & 720 & 384 \\
Retail L2 & 7 & 810 & 765 \\
Retail L3 & 8 & 624 & 537 \\
Banking L1 & 4 & 336 & 178 \\
Banking L2 & 5 & 500 & 387 \\
Banking L3 & 6 & 600 & 551 \\
Pitch L1 & 4 & 216 & 96 \\
Pitch L2 & 6 & 324 & 144 \\
Pitch L3 & 7 & 378 & 168 \\
\midrule
\textbf{Total} & \textbf{53} & \textbf{4,508} & \textbf{3,210} \\
\bottomrule
\end{tabular}
\vspace{-0.5cm}
\end{wraptable}
\paragraph{Banking.}
Banking is a case-review environment covering three core decision types: authorizing pending card transactions, reviewing credit-limit increases, and screening outbound transfers. 
Harder variants compose these decisions into ranked alternatives or multi-line payment batches, requiring selection of the first admissible option or a verdict for each line. 
Each task operates on an account and case in a database-backed environment, with hidden policies governing verification requirements, spending and travel limits, account standing, requested-limit caps, merchant restrictions, and payee-age thresholds.

\paragraph{Pitch.}
Pitch is a single-turn generation environment in which each task provides a factual product sheet and asks the agent to produce a 40--80 word sales pitch. The hidden state is the active customer preference profile, and a graded verifier rewards coverage of positively valued attributes while penalizing undesirable attributes and unsupported claims. Because the appropriate emphasis always depends on this latent profile, all Pitch tasks are Hidden-Dependent.

\paragraph{Difficulty design.}
The three difficulty tiers vary how environment knowledge changes over time and how task complexity is expressed across domains. \emph{L1 (acquisition)} introduces previously unseen states, using mostly single-policy Retail and Banking tasks and non-repeating Pitch profiles. \emph{L2 (revision and composition)} revises earlier states, adds multi-policy Retail requests and ranked or batched Banking decisions, and introduces recurring Pitch profiles over a broader set of attributes. \emph{L3 (non-monotonic evolution)} further introduces reversals, removals, and re-entry while retaining L2-style Retail and Banking composition; Pitch likewise cycles back to previously seen profiles.

\paragraph{Reward metric.}
ServeLearnBench uses domain-specific reward functions. Retail rewards correct database transitions or refusal reasons; Banking rewards correct verdicts and resulting states; and Pitch uses a structured rubric that rewards preference coverage while penalizing undesirable attributes and unsupported claims. Appendix~\ref{app:task-evaluation} provides the full evaluation details.

\paragraph{Reference baselines.}
We include two non-learning baselines to contextualize performance. \emph{Blind} uses only learner-visible task information and retains no serving experience, measuring performance without adaptation. \emph{Oracle} additionally receives the active hidden policy for each window, measuring performance when hidden-state inference is removed while all other task demands remain unchanged.

%% file: tex/04_experimental_setup.tex
\section{Evaluation}
\label{sec:evaluation}
\label{sec:results}

We evaluate ServeLearnBench by crossing models with learning harnesses across all scenario--difficulty settings.
We first describe the experimental setting (Section~\ref{sec:setup}), then present the main evaluation matrix (Section~\ref{sec:main-evaluation}), and finally organize the analysis around three questions: how much learnable headroom the benchmark creates, how much of that headroom a streaming learner captures, and whether adaptation harms fully specified cases.

\subsection{Experimental Setting}
\label{sec:setup}

\paragraph{Model.}
The main evaluation compares six models: GPT-5.6 Terra~\citep{openai2026gpt56terra}, Opus 5~\citep{anthropic2026opus5}, Kimi K3~\citep{kimiteam2026k3}, GLM-5.3~\citep{zai2026glm53}, DeepSeek V4.1 Flash~\citep{deepseekai2026v41flash}, and GLM-5.3 Flash~\citep{zai2026glm53flash}.
GPT-5.6 Terra and Opus 5 are accessed through the OpenAI and Anthropic APIs, respectively; the other four models are accessed through Fireworks.

\paragraph{Harnesses.}
The main comparison includes Blind and Oracle references alongside five learning harnesses: RAG~\citep{lewis2020rag}, Mem0~\citep{chhikara2025mem0}, SkillOpt~\citep{yang2026skillopt}, Continual Harness (CH)~\citep{karten2026continualharness}, and Prime~\citep{karten2026primeagent}. Crossing six models with the five harnesses yields 30 possible model--harness pairs across nine scenario--difficulty settings; 28 were evaluated, with two omitted due to cost (Table~\ref{tab:main-results}). We use identical task streams and scoring across methods while preserving each harness's native runtime, update schedule, replay access, learning computation, and state capacity. Implementation details and operating conditions are provided in Appendix~\ref{app:learning-harnesses}, with model and evaluation settings in Appendix~\ref{app:additional-experimental-setting}.

\par

%% file: tex/05_results.tex
\input{tex/main_results_table}

\subsection{Main Evaluation Results}
\label{sec:main-evaluation}

Table~\ref{tab:main-results} compares the performance of different model--harness combinations on ServeLearnBench.
Across six models, Oracle achieves an average Hidden score of 95.4, compared with 14.1 for Blind, showing that access to environment knowledge substantially improves performance.
However, knowing that knowledge does not always lead to better behavior: on Pitch L1, Opus 5 scores 90.8 with Oracle, below the 99.7 achieved with CH.
When informed of the customer's dislikes, Oracle sometimes produces misleading claims to appeal to those preferences.
For example, it describes an already-trending product as one to buy ``before word spreads,'' contradicting the product sheet and incurring penalties.
CH instead learns to mention the preferred attributes and omit the others; both achieve full preference coverage, but CH avoids nearly all such penalties.

\begin{wrapfigure}{R}{0.34\textwidth}
  \centering
  \vspace{-0.2cm}
  \includegraphics[width=\linewidth]{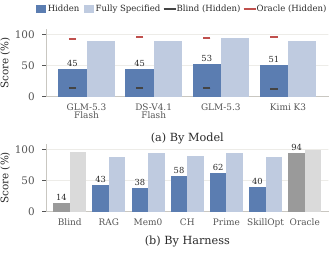}
  \caption{\textbf{Average performance} 
  \textbf{(a)} by model and \textbf{(b)} by harness.}
  \label{fig:avg-performance}
  \vspace{-0.4cm}
\end{wrapfigure}
Among the learning harnesses, Prime and CH generally achieve stronger Hidden performance than RAG, Mem0, and SkillOpt, although substantial gaps to Oracle remain.
Figure~\ref{fig:avg-performance} summarizes this pattern across the four open-weight models: Prime and CH average 61.7 and 57.6, respectively, compared with 38.3--43.4 for the other three harnesses.
On Kimi K3, for example, Prime reaches 61.7, versus 45.0 for RAG and 37.7 for Mem0.
The ranking is not uniform across models: CH performs best on GLM-5.3, scoring 69.8 against Prime's 63.7, while Prime leads on the other three open-weight models.
Higher Hidden performance can also come with losses on Fully Specified tasks, so improvements in learning must be considered alongside preservation of existing capabilities (Section~\ref{sec:results-interference}).

\vspace{-0.2cm}
\subsection{Cost Comparison}
\label{sec:results-run-cost}

\begin{wrapfigure}{R}{0.46\textwidth}
  \centering
  \vspace{-0.3cm}
  \includegraphics[width=\linewidth]{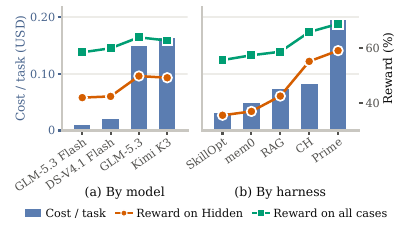}
\caption{\textbf{Cost vs.\ performance.}
Average API cost per task and reward, aggregated \textbf{(a)} by model and \textbf{(b)} by learning harness. Bars show cost, including learning; lines show Hidden and overall reward, averaged equally across the nine settings. Blind and Oracle are excluded.}
  \label{fig:cost-merged}
\end{wrapfigure}

\paragraph{Performance--resource frontiers.}
\label{sec:results-frontier}
Accuracy alone does not determine whether a deployed learner is useful: adaptation changes both the reasoning emitted by the model and the auxiliary calls made by its harness.
Figure~\ref{fig:cost-frontier} compares reward on Hidden cases with the tokens consumed and the API cost per held-out test task; tokens count every input token of every model call, cached or not, plus the output tokens.
Figure~\ref{fig:cost-frontier} covers the 28 model--harness pairs of Table~\ref{tab:main-results} (252 learning lanes) and uses the table's held-out test Avg.~H as its reward axis. The grouped cost comparisons use the four open-weight models (180 lanes), with costs computed from archived runs and rewards averaged equally across the nine settings. Figure~\ref{fig:cost-merged} combines serving and test tasks within each setting, whereas Figure~\ref{fig:cost-correctness} reports the two phases separately.
The dashboard labels \emph{adapt} and \emph{general} correspond to Hidden and Fully Specified cases, respectively.
API cost accounts for uncached input, cache reads, cache writes, and output at each provider's official prices, so token efficiency and monetary efficiency can induce different rankings.

\paragraph{Cost per task against reward.}
\label{sec:results-serving-cost}
Figure~\ref{fig:cost-merged} compares average per-task API cost with reward, grouped by acting model and learning harness. Cost includes both serving and held-out test execution, including any serving-side learning computation, and reward is averaged across the same tasks and then equally across the nine settings. Across harnesses, higher cost generally accompanies higher Hidden reward: Prime achieves the highest reward (59.0) at \$0.195 per task, while CH reaches 55.1 at less than half the cost (\$0.083). RAG, Mem0, and SkillOpt are cheaper but trail CH by 13--20 points. Across models, GLM-5.3 and Kimi K3 achieve the highest Hidden reward (49.8 and 49.2), but at 14--16$\times$ the per-task cost of GLM-5.3 Flash (\$0.010), which reaches 42.0. Appendix~\ref{app:cost-breakdown} provides phase-level and learning-cost breakdowns.

\FloatBarrier
\vspace{-0.2cm}
\section{Analysis}
\label{sec:analysis}
\vspace{-0.1cm}

The comparison above ranks harnesses and models by their final scores; this section asks how those scores come about: how quickly a harness picks up a changed policy, what adaptation does to the cases that needed no learning, and whether the way a harness explores and retains predicts its reward.

\vspace{-0.1cm}
\subsection{How Fast Do Harnesses Learn a New Policy?}
\label{sec:results-learning-speed}

\begin{figure*}[t]
  \centering
  \includegraphics[width=0.94\textwidth]{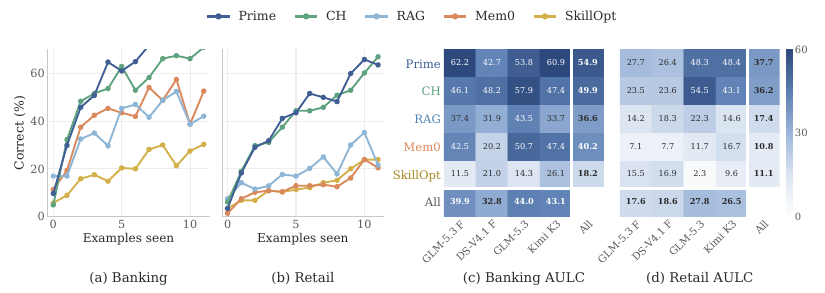}
  \caption{\textbf{Learning speed for new policies.}
  \textbf{(a, b)} Correctness on Hidden serving tickets as a function of the number of prior tickets from the same policy, pooled across four open-weight models for Banking and Retail, respectively.
  \textbf{(c, d)} AULC$_{10}$ (\%) by model and harness.}
  \label{fig:learning-speed}
\end{figure*}

Final performance does not reveal how quickly a harness adapts to a new policy. We therefore track each policy from the window in which it becomes active until it changes again, and measure correctness on the $k$-th Hidden serving example governed by that policy. We perform this analysis on Banking and Retail, where Hidden tickets can be attributed to explicit rules and states; Pitch is excluded because its latent preference profiles do not decompose into such rules. The analysis uses the recorded serving streams of four open-weight models across all five learning harnesses, covering 31 Banking and 37 Retail policies (Appendix~\ref{app:learning-speed}).

Let $a_k$ denote mean correctness after exactly $k$ previous examples of the same policy. We summarize learning speed using $\mathrm{AULC}_K = \frac{1}{K}\sum_{k=0}^{K-1} a_k$, with $K=10$. AULC$_{10}$ measures average correctness over the first ten encounters with a new policy, rewarding both rapid and high-quality learning without requiring a threshold for when a policy is considered learned.

Figure~\ref{fig:learning-speed} reveals substantial differences in learning speed across harnesses. Prime and CH adapt fastest in both domains, reaching AULC$_{10}$ scores of 54.9 and 49.9 in Banking and 37.7 and 36.2 in Retail; after five examples of a new Banking policy, both answer roughly 60\% of subsequent tickets correctly. Mem0 and RAG learn more slowly, particularly in Retail, while SkillOpt is slowest in both domains, with AULC$_{10}$ scores of 18.2 in Banking and 11.1 in Retail. Learning speed also varies substantially across acting models: GLM-5.3 and Kimi K3 achieve the highest pooled AULC$_{10}$ in Banking, and Retail is consistently harder to learn across models. Appendix~\ref{app:learning-harnesses} discusses harness-specific learning mechanisms, and Appendix~\ref{app:learn-vs-drop} compares acquiring a policy with dropping it once the corresponding rule is withdrawn.

\needlines{16}
\subsection{Does Adaptation Harm Fully Specified Cases?}
\label{sec:results-interference}

\begin{wrapfigure}{r}{0.46\textwidth}
  \centering
  \vspace{-0.5cm}
  \includegraphics[width=\linewidth]{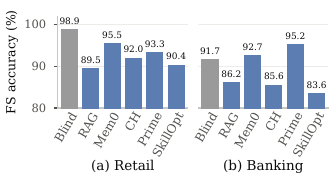}
  \caption{\textbf{Fully Specified accuracy with and without adaptation.}
  Mean over four open-weight models and three difficulty levels.}
  \vspace{-0.4cm}
  \label{fig:fs-method-comparison}
\end{wrapfigure}
Fully Specified scores reveal effects that Hidden improvements alone cannot capture.
Figure~\ref{fig:fs-method-comparison} averages them over the four open-weight models: every learning harness loses Fully Specified accuracy against Blind in Retail (98.9 for Blind against 89.5--95.5), and all but Prime and Mem0 lose in Banking (91.7 against 83.6--86.2, Prime 95.2, Mem0 92.7).
On GLM-5.3 Flash, CH raises average Hidden performance from Blind's 14.6 to 49.7, while Fully Specified performance falls from 96.7 to 82.8.
Prime attains higher Hidden performance, 61.1, while retaining 95.2 on Fully Specified cases.
Pitch contributes only to Hidden averages because the operative customer preferences remain latent.

\subsection{Does Exploration Predict Adaptation?}
\label{sec:results-exploration}

\begin{figure*}[t]
  \centering
  \includegraphics[width=0.94\textwidth]{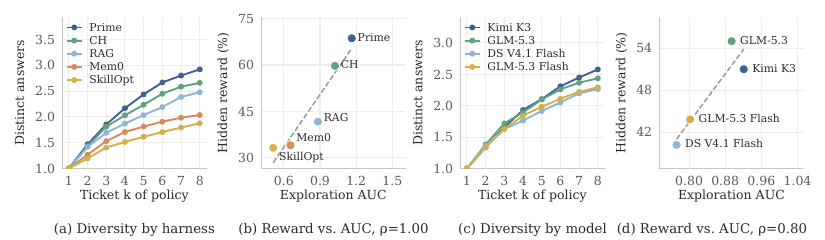}
  \caption{\textbf{Exploration across models and harnesses.}
  \textbf{(a, c)} Mean number of distinct answers after $k$ Hidden serving examples of the same policy, aggregated by harness and model.
  \textbf{(b, d)} Exploration AUC versus mean Hidden test reward across the six Retail and Banking settings; $\rho$ denotes Spearman rank correlation.
  Per-domain results are provided in Appendix~\ref{app:exploration}.}
  \label{fig:exploration}
  \vspace{-0.3cm}
\end{figure*}

\textbf{Exploration.} A harness receives only outcome feedback after each serving ticket, so discovering a changed policy often requires trying behavior beyond what the visible specification suggests. We therefore measure \emph{exploration} by how many distinct answers a harness produces while encountering the same hidden policy, regardless of correctness. We analyze Retail policies whose correct behavior is a refusal and Banking tickets governed by a single rule; Appendix~\ref{app:exploration} provides more details.

For a policy, let $c_k$ denote the number of distinct answers produced over its first $k$ Hidden serving tickets. We summarize exploration over the first $K$ tickets as $\mathrm{AUC}_{\mathrm{E}}=\frac{1}{K}\sum_{k=1}^{K}(c_k-1)$, with $K=8$. A harness that always repeats the same answer has $\mathrm{AUC}_{\mathrm{E}}=0$, while larger values indicate greater behavioral exploration. Results are robust to $K\in[4,12]$ (Appendix~\ref{app:exploration}).

Figure~\ref{fig:exploration} shows a strong association between exploration and adaptation performance. Prime, CH, RAG, Mem0, and SkillOpt have exploration AUCs of 1.16, 1.02, 0.88, 0.65, and 0.51, respectively, exactly matching their ordering in mean Hidden reward and yielding $\rho=1.00$. The relationship persists at the finer harness--domain--tier level ($\rho=0.75$; Appendix~\ref{app:exploration}). The gap is especially pronounced in Retail, where Prime tries substantially more distinct answers while Mem0 and SkillOpt often continue repeating the visible-specification behavior. Across models, the same relationship holds but variation is smaller: exploration spans 0.77--0.92 across models, compared with 0.51--1.16 across harnesses. These results suggest that effective exploration is primarily a property of the learning harness and is strongly associated with successful continual adaptation. 

\begin{wrapfigure}{r}{0.46\textwidth}
  \centering
  \vspace{-0.2cm}
  \includegraphics[width=\linewidth]{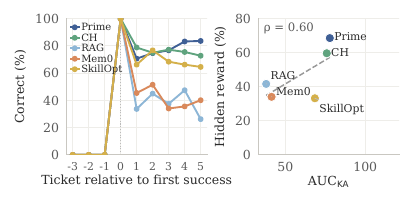}
\caption{\textbf{Knowledge acquisition after first success.}
Left: correctness on Hidden serving tickets aligned at the first correct response (ticket 0), averaged across Retail and Banking.
Right: mean correctness over the next five tickets ($\mathrm{AUC}_{\mathrm{KA}}$) versus mean Hidden test reward; $\rho$ denotes Spearman rank correlation.}
\vspace{-0.3cm}
  \label{fig:consolidation}
\end{wrapfigure}
\textbf{Knowledge Acquisition.}
A first success does not necessarily mean that an agent has learned the underlying policy. We therefore measure whether that success becomes reusable knowledge by aligning each policy at its first correct ticket and tracking correctness over the next $K$ occurrences. Letting $r_j$ denote correctness on the $j$-th subsequent ticket, we define the \emph{knowledge-acquisition AUC} as $\mathrm{AUC}_{\mathrm{KA}}=\frac{1}{K}\sum_{j=1}^{K} r_j$, and use $K=5$ throughout our analysis (Appendix~\ref{app:exploration}).

Figure~\ref{fig:consolidation} shows large differences across harnesses. Prime and CH achieve $\mathrm{AUC}_{\mathrm{KA}}$ values of 78\% and 76\%, followed by SkillOpt at 68\%, while Mem0 and RAG reach only 41\% and 38\%. The post-success curves are largely flat, suggesting that successful behavior either becomes reusable almost immediately or remains unreliable over subsequent encounters. Knowledge acquisition also differs from exploration: its rank correlation with Hidden reward is only $0.60$. SkillOpt performs well once it discovers the correct behavior but struggles to find it in the first place, whereas RAG often discovers the correct behavior but fails to reuse it reliably. 

%% file: tex/main_results_table.tex
% Generated table; edit the pinned source, not this file.
% Source commit: 149446dfa8c08b073cd8daeadcf21f2d979676bb+dirty; snapshot: 2026-09-25 16:32
\begingroup
\begin{table}[t!]
\centering
\scriptsize
\setlength{\tabcolsep}{1.5pt}
\renewcommand{\arraystretch}{0.9}
\caption{Performance comparison of five harnesses across six models on ServeLearnBench.}
\label{tab:main-results}
\begin{tabular*}{\linewidth}{@{\extracolsep{\fill}}cl!{\vrule width 0.35pt}*{3}{c}!{\vrule width 0.35pt}*{3}{c}!{\vrule width 0.35pt}*{3}{c}!{\vrule width 0.35pt}*{3}{c}!{\vrule width 0.35pt}*{3}{c}!{\vrule width 0.35pt}*{2}{c}!{\vrule width 0.35pt}c@{}}
\toprule
\multicolumn{1}{c}{\multirow{3}{*}[-1.0ex]{Model}} &
\multicolumn{1}{c}{\multirow{3}{*}[-1.0ex]{Harness}} &
\multicolumn{6}{c}{Retail} & \multicolumn{6}{c}{Banking} & \multicolumn{3}{c}{Pitch} &
\multicolumn{2}{c}{Avg.} &
\multicolumn{1}{c}{\multirow{3}{*}[-1.0ex]{Cost}} \\
\cmidrule(lr){3-8}\cmidrule(lr){9-14}\cmidrule(lr){15-17}\cmidrule(lr){18-19}
 & & \multicolumn{3}{c}{\scriptsize Hidden} & \multicolumn{3}{c}{\scriptsize Fully Specified}
 & \multicolumn{3}{c}{\scriptsize Hidden} & \multicolumn{3}{c}{\scriptsize Fully Specified}
 & \multicolumn{3}{c}{\scriptsize Hidden} & H & F & \\
\cmidrule(lr){3-5}\cmidrule(lr){6-8}
\cmidrule(lr){9-11}\cmidrule(lr){12-14}
\cmidrule(lr){15-17}
 & & L1 & L2 & L3 & L1 & L2 & L3
 & L1 & L2 & L3 & L1 & L2 & L3
 & L1 & L2 & L3 & & & \\
\midrule
\rowcolor{gray!12}
\cellcolor{white} & Oracle
 & 100.0 & 96.0 & 95.5 & 100.0 & 100.0 & 99.7
 & 100.0 & 98.9 & 98.7 & 100.0 & 100.0 & 99.2
 & 89.2 & 92.1 & 89.5 & 95.5 & 99.8 & 102 \\*
\rowcolor{gray!12}
\cellcolor{white} & Blind
 & 1.7 & 0.0 & 2.2 & 100.0 & 100.0 & 99.0
 & 1.2 & 14.0 & 16.0 & 95.7 & 96.5 & 96.6
 & 31.7 & 25.9 & 18.3 & 12.3 & 98.0 & 120 \\*
 & RAG
 & 65.4 & 18.8 & 15.7 & 86.8 & 90.8 & 95.9
 & 69.0 & 60.8 & 35.8 & 91.5 & 85.1 & 86.1
 & 47.9 & 51.0 & 40.8 & 45.0 & 89.4 & 1,073 \\*
 & Mem0
 & 20.4 & 14.4 & 23.8 & 94.4 & 91.6 & 94.6
 & 69.0 & 55.4 & 35.5 & 97.9 & 93.5 & 87.8
 & 34.4 & 48.1 & 38.0 & 37.7 & 93.3 & 829 \\*
 & CH
 & 100.0 & 54.2 & 33.6 & 99.3 & 82.0 & 83.1
 & 51.2 & 54.3 & 51.1 & 69.1 & 87.1 & 84.9
 & 68.4 & 56.0 & 51.4 & 57.8 & 84.2 & 1,335 \\*
 & Prime
 & 87.1 & 61.6 & 52.0 & 95.8 & 89.1 & 93.6
 & 97.6 & 79.0 & 67.4 & 98.9 & 91.5 & 94.1
 & 64.0 & 16.9 & 29.9 & 61.7 & 93.8 & 2,479 \\*
\cellcolor{white}\multirow{-7}{*}{\shortstack{Kimi K3\\{\scriptsize (2.8T-A104B)}}} & SkillOpt
 & 38.8 & 4.8 & 37.7 & 81.9 & 87.9 & 86.6
 & 46.4 & 68.8 & 59.7 & 79.8 & 88.6 & 81.1
 & 65.3 & 74.2 & 64.6 & 51.1 & 84.3 & 605 \\*
\midrule
\rowcolor{gray!12}
\cellcolor{white} & Oracle
 & 100.0 & 95.2 & 92.4 & 99.3 & 97.9 & 97.8
 & 100.0 & 100.0 & 99.7 & 100.0 & 100.0 & 99.2
 & 79.4 & 90.1 & 85.7 & 93.6 & 99.0 & 29.1 \\*
\rowcolor{gray!12}
\cellcolor{white} & Blind
 & 13.3 & 0.0 & 1.8 & 100.0 & 99.6 & 95.9
 & 0.0 & 16.7 & 15.7 & 91.5 & 92.5 & 92.4
 & 32.6 & 25.8 & 19.5 & 13.9 & 95.3 & 40.2 \\*
 & RAG
 & 62.9 & 30.0 & 15.2 & 91.7 & 87.9 & 84.4
 & 58.3 & 64.0 & 55.6 & 88.3 & 89.6 & 84.0
 & 45.0 & 44.5 & 37.4 & 45.9 & 87.6 & 932 \\*
 & Mem0
 & 20.8 & 12.4 & 17.5 & 98.6 & 96.2 & 95.5
 & 66.7 & 69.4 & 58.1 & 98.9 & 94.0 & 92.4
 & 51.3 & 52.0 & 45.0 & 43.7 & 95.9 & 565 \\*
 & CH
 & 100.0 & 72.8 & 59.6 & 99.3 & 93.7 & 98.4
 & 88.1 & 89.2 & 59.4 & 93.6 & 98.0 & 97.1
 & 76.3 & 41.9 & 40.9 & 69.8 & 96.7 & 1,077 \\*
 & Prime
 & 90.8 & 61.4 & 39.5 & 97.2 & 87.9 & 98.1
 & 94.0 & 69.4 & 78.9 & 97.9 & 96.5 & 96.2
 & 73.8 & 41.4 & 24.0 & 63.7 & 95.6 & 2,930 \\*
\cellcolor{white}\multirow{-7}{*}{\shortstack{GLM-5.3\\{\scriptsize (743B-A40B)}}} & SkillOpt
 & 69.2 & 0.0 & 4.9 & 91.7 & 92.9 & 96.5
 & 81.0 & 29.0 & 31.3 & 90.4 & 90.5 & 86.1
 & 49.7 & 64.7 & 25.6 & 39.5 & 91.4 & 251 \\*
\midrule
\rowcolor{gray!12}
\cellcolor{white} & Oracle
 & 100.0 & 99.6 & 93.7 & 100.0 & 99.6 & 100.0
 & 100.0 & 100.0 & 100.0 & 100.0 & 100.0 & 99.6
 & 91.4 & 90.6 & 85.1 & 95.6 & 99.9 & 7.4 \\*
\rowcolor{gray!12}
\cellcolor{white} & Blind
 & 1.2 & 0.2 & 0.9 & 100.0 & 99.2 & 99.0
 & 0.0 & 19.4 & 15.7 & 88.3 & 82.1 & 78.2
 & 37.8 & 32.4 & 21.9 & 14.4 & 91.1 & 11.0 \\*
 & RAG
 & 51.2 & 23.6 & 13.9 & 91.7 & 80.8 & 80.3
 & 52.4 & 48.4 & 32.6 & 85.1 & 77.1 & 84.9
 & 68.7 & 61.3 & 46.7 & 44.3 & 83.3 & 192 \\*
 & Mem0
 & 18.8 & 4.2 & 13.9 & 95.8 & 90.4 & 98.7
 & 44.0 & 36.0 & 28.1 & 95.7 & 96.5 & 88.7
 & 59.8 & 49.0 & 47.8 & 33.5 & 94.3 & 75.9 \\*
 & CH
 & 50.0 & 24.1 & 22.4 & 95.8 & 77.8 & 99.0
 & 83.3 & 61.8 & 73.8 & 89.4 & 91.0 & 95.0
 & 54.4 & 52.0 & 58.1 & 53.3 & 91.3 & 52.0 \\*
 & Prime
 & 80.0 & 64.1 & 25.1 & 95.1 & 86.2 & 92.0
 & 71.4 & 70.4 & 64.9 & 97.9 & 89.1 & 94.1
 & 76.1 & 50.9 & 40.2 & 60.3 & 92.4 & 429 \\*
\cellcolor{white}\multirow{-7}{*}{\shortstack{DeepSeek V4.1\\Flash\\{\scriptsize (552B-A16B)}}} & SkillOpt
 & 23.8 & 12.0 & 22.9 & 93.8 & 85.8 & 86.6
 & 19.0 & 37.6 & 31.9 & 85.1 & 95.5 & 75.6
 & 49.1 & 53.6 & 48.6 & 33.2 & 87.1 & 39.8 \\*
\midrule
\rowcolor{gray!12}
\cellcolor{white} & Oracle
 & 98.8 & 93.5 & 89.2 & 100.0 & 95.4 & 95.5
 & 100.0 & 98.9 & 99.4 & 100.0 & 100.0 & 100.0
 & 83.9 & 89.9 & 83.9 & 93.1 & 98.5 & 5.0 \\*
\rowcolor{gray!12}
\cellcolor{white} & Blind
 & 14.6 & 0.4 & 4.5 & 100.0 & 99.6 & 94.3
 & 0.0 & 15.6 & 17.3 & 95.7 & 95.0 & 95.4
 & 33.8 & 25.4 & 20.2 & 14.6 & 96.7 & 6.8 \\*
 & RAG
 & 45.4 & 14.6 & 14.3 & 100.0 & 88.7 & 95.2
 & 69.0 & 54.3 & 29.7 & 95.7 & 85.1 & 81.9
 & 50.1 & 44.5 & 22.3 & 38.2 & 91.1 & 56.3 \\*
 & Mem0
 & 18.3 & 4.6 & 13.5 & 99.3 & 97.9 & 92.7
 & 67.9 & 57.0 & 49.2 & 87.2 & 90.0 & 89.1
 & 51.9 & 43.2 & 39.7 & 38.4 & 92.7 & 48.8 \\*
 & CH
 & 75.8 & 29.7 & 28.7 & 97.9 & 85.4 & 92.0
 & 53.6 & 47.3 & 69.3 & 63.8 & 76.6 & 81.1
 & 48.2 & 57.6 & 36.7 & 49.7 & 82.8 & 88.5 \\*
 & Prime
 & 76.7 & 32.1 & 26.9 & 99.3 & 88.3 & 97.5
 & 94.0 & 84.4 & 78.6 & 98.9 & 93.0 & 94.5
 & 70.0 & 48.8 & 38.6 & 61.1 & 95.2 & 172 \\*
\cellcolor{white}\multirow{-7}{*}{\shortstack{GLM-5.3\\Flash\\{\scriptsize (320B-A18B)}}} & SkillOpt
 & 20.8 & 31.7 & 22.0 & 100.0 & 87.4 & 93.3
 & 32.1 & 34.9 & 38.0 & 80.9 & 70.1 & 79.0
 & 40.4 & 52.2 & 46.4 & 35.4 & 85.1 & 35.1 \\*
\midrule
\rowcolor{gray!12}
\cellcolor{white} & Oracle
 & 100.0 & 99.6 & 99.6 & 100.0 & 100.0 & 99.7
 & 100.0 & 100.0 & 100.0 & 100.0 & 100.0 & 100.0
 & 90.8 & 91.0 & 93.3 & 97.1 & 100.0 & 131 \\*
\rowcolor{gray!12}
\cellcolor{white} & Blind
 & 0.0 & 1.9 & 1.8 & 100.0 & 97.9 & 99.7
 & 0.0 & 23.7 & 18.8 & 69.1 & 52.7 & 58.8
 & 36.3 & 30.9 & 17.3 & 14.5 & 79.7 & 207 \\*
 & RAG
 & 64.2 & 23.8 & 17.5 & 88.9 & 84.5 & 87.6
 & 66.7 & 42.5 & 25.9 & 94.7 & 86.6 & 86.6
 & 48.6 & 43.0 & 35.7 & 40.9 & 88.1 & 2,602 \\*
 & Mem0
 & 27.9 & 14.4 & 15.7 & 90.3 & 80.3 & 85.4
 & 58.3 & 67.2 & 55.0 & 97.9 & 94.0 & 93.7
 & 68.9 & 52.1 & 70.9 & 47.8 & 90.3 & 2,003 \\*
 & CH
 & 97.9 & 51.5 & 30.5 & 99.3 & 93.7 & 97.8
 & 82.1 & 80.1 & 76.0 & 96.8 & 95.0 & 94.5
 & 99.7 & 83.6 & 58.9 & 73.4 & 96.2 & 2,449 \\*
\cellcolor{white}\multirow{-6}{*}{Opus 5} & SkillOpt
 & 69.2 & 1.1 & 17.0 & 97.9 & 93.3 & 89.8
 & 57.1 & 67.7 & 49.5 & 94.7 & 87.6 & 92.9
 & 57.6 & 72.6 & 61.9 & 50.4 & 92.7 & 629 \\*
\midrule
\rowcolor{gray!12}
\cellcolor{white} & Oracle
 & 99.6 & 95.1 & 91.9 & 99.3 & 98.7 & 98.1
 & 100.0 & 98.9 & 98.1 & 96.8 & 99.0 & 100.0
 & 98.5 & 98.1 & 98.6 & 97.6 & 98.6 & 35.6 \\*
\rowcolor{gray!12}
\cellcolor{white} & Blind
 & 0.0 & 0.0 & 0.0 & 100.0 & 97.9 & 99.4
 & 0.0 & 14.5 & 16.6 & 97.9 & 95.0 & 96.6
 & 43.9 & 35.0 & 25.5 & 15.1 & 97.8 & 39.6 \\*
 & RAG
 & 27.9 & 11.4 & 13.9 & 97.9 & 92.9 & 94.3
 & 52.4 & 30.1 & 23.6 & 91.5 & 87.1 & 84.9
 & 59.7 & 55.9 & 39.6 & 34.9 & 91.4 & 496 \\*
 & Mem0
 & 0.8 & 0.2 & 6.3 & 100.0 & 97.9 & 98.4
 & 56.0 & 15.1 & 30.4 & 100.0 & 88.6 & 91.6
 & 72.2 & 65.5 & 38.9 & 31.7 & 96.1 & 206 \\*
 & CH
 & 53.3 & 23.4 & 30.5 & 96.5 & 82.4 & 98.7
 & 86.9 & 57.0 & 69.6 & 95.7 & 96.0 & 95.8
 & 63.7 & 63.6 & 54.6 & 55.8 & 94.2 & 239 \\*
\cellcolor{white}\multirow{-6}{*}{\shortstack{GPT-5.6\\Terra}} & SkillOpt
 & 34.2 & 0.2 & 26.9 & 100.0 & 87.0 & 94.3
 & 19.0 & 14.0 & 42.8 & 100.0 & 99.5 & 86.6
 & 55.0 & 42.7 & 31.2 & 29.6 & 94.6 & 146 \\*
\bottomrule
\end{tabular*}
\end{table}
\endgroup

%% file: tex/07_conclusion.tex
\vspace{-0.2cm}
\section{Conclusion}
\vspace{-0.1cm}

We introduce ServeLearnBench, a benchmark for continual learning from serving experience in hidden, evolving environments. Across three domains and nine scenarios, our evaluation shows that substantial gaps remain in acquiring and revising hidden environment knowledge, even when the underlying tasks are well within model capabilities once the relevant policy is disclosed. Continual adaptation also comes with significant computational cost and can degrade behavior that was already correct. Our analyses further identify exploration and knowledge acquisition as distinct bottlenecks: agents must not only discover useful behavior, but also convert successful experience into reusable knowledge. We hope ServeLearnBench provides a systematic testbed for diagnosing these challenges and tracking progress toward agents that learn continually and reliably from deployment experience.

%% file: tex/a_overview.tex
\section{Appendix Overview}
\label{app:overview}

\begin{itemize}[leftmargin=1.4em, itemsep=3pt, topsep=2pt]
\item \textbf{Appendix~\ref{app:additional-experimental-setting}: \nameref{app:additional-experimental-setting}.} The setting behind Section~\ref{sec:evaluation}.
  \begin{itemize}[leftmargin=1.6em, itemsep=0pt, topsep=1pt]
  \item \ref{app:models-budgets}\quad \nameref{app:models-budgets}
  \item \ref{app:learning-harnesses}\quad \nameref{app:learning-harnesses}
  \item \ref{app:evaluation-protocol}\quad \nameref{app:evaluation-protocol}
  \item \ref{app:data-reporting}\quad \nameref{app:data-reporting}
  \end{itemize}
\item \textbf{Appendix~\ref{app:benchmark-details}: \nameref{app:benchmark-details}.} Interfaces, tiers, hidden rules, and verification for the nine scenarios of Section~\ref{sec:benchmark}.
  \begin{itemize}[leftmargin=1.6em, itemsep=0pt, topsep=1pt]
  \item \ref{app:interfaces}\quad \nameref{app:interfaces}
  \item \ref{app:tiers}\quad \nameref{app:tiers}
  \item \ref{app:retail-rules}\quad \nameref{app:retail-rules}
  \item \ref{app:banking-rules}\quad \nameref{app:banking-rules}
  \item \ref{app:pitch-rules}\quad \nameref{app:pitch-rules}
  \item \ref{app:task-evaluation}\quad \nameref{app:task-evaluation}
  \end{itemize}
\item \textbf{Appendix~\ref{app:additional-evaluation}: \nameref{app:additional-evaluation}.} Methods and extra results behind the analyses of Section~\ref{sec:analysis}.
  \begin{itemize}[leftmargin=1.6em, itemsep=0pt, topsep=1pt]
  \item \ref{app:learning-speed}\quad \nameref{app:learning-speed}
  \item \ref{app:learn-vs-drop}\quad \nameref{app:learn-vs-drop}
  \item \ref{app:exploration}\quad \nameref{app:exploration}
  \item \ref{app:cost-breakdown}\quad \nameref{app:cost-breakdown}
  \item \ref{app:training-based-methods}\quad \nameref{app:training-based-methods}
  \item \ref{app:fs-reference-performance}\quad \nameref{app:fs-reference-performance}
  \end{itemize}
\end{itemize}

%% file: tex/a_additional_experimental_setting.tex
\section{Additional Experimental Setting}
\label{app:additional-experimental-setting}
\small

\subsection{Model and Budgets}
\label{app:models-budgets}
All reported cells use the unified high-effort regime.
Within each model block, Blind, Oracle, and the learning harnesses use the same model so that differences among harnesses reflect access to environment knowledge and serving history rather than base-model capability.
The provider endpoints expose different reasoning histories: Terra returns no chain of thought and Opus returns a summary, while the Fireworks trajectories retain the full returned reasoning.
Harnesses that learn from trajectories therefore receive different amounts of reasoning information across these providers.
Pitch uses a fixed GLM-5.3 Flash judge with high reasoning effort to produce structured rubric labels, which deterministic code maps to the continuous rating specified in Appendix~\ref{app:task-evaluation}.
The same judge scores all models, including GLM-5.3 Flash itself.
The shared loop supplies tools through the provider's native tool-calling interface, returns observations as tool-result messages, and exposes \texttt{finish} as a tool.
It permits one tool call per turn; a response containing multiple calls is rejected without executing any of them.
Each scored item has one attempt. The shared loop imposes a budget of 50 environment tool calls with terminal actions exempt, a 100-step runner guard, and a 256k output-token cap.
Assistant messages, including reasoning, are replayed as returned by the provider.
Prime retains its own action interface, with persistent harness state and a fresh session for each ticket.
Test-side tasks remain excluded from all learner updates, validation, and candidate selection.
Infrastructure failures trigger task reruns rather than scored failures; three consecutive infrastructure failures stop the lane.
Model context overflow is scored on the task outcome as it stands.

\subsection{Learning Harnesses}
\label{app:learning-harnesses}

The five learning harnesses are complete systems of very different size and design: Prime is an agent runtime with its own sandbox and code interpreter, SkillOpt is a prompt optimizer built for batched offline training, Mem0 and CH are memory layers with their own extraction or refinement models, and RAG is a retrieval index with no learning-side model calls.
Aligning them to one update cadence, one state budget or one learning-side compute budget would change the methods themselves, so we did not attempt it.
The comparison uses the same task streams and held-out test sets, the same model and reasoning effort within each model block, the same per-task budget (50 tool calls, 100 turns), and score-only outcome feedback. Hidden-policy descriptions and corrected actions are withheld from learning harnesses, and held-out test tasks and feedback are excluded from learning (Appendix~\ref{app:evaluation-protocol}).
Update schedules, replay access, state capacity, and learning-side computation follow each method's design and the adaptations described below. Comparisons therefore characterize complete systems under their disclosed operating conditions; they do not isolate harness effects under identical information and resource budgets.

\paragraph{Acting agent.}
RAG, Mem0, SkillOpt and CH share one ReAct acting loop with one tool call per turn: RAG injects the retrieved cases and SkillOpt its skill document into the system prompt, Mem0 adds two memory tools, and CH adds the Continual Harness tools for memories, skills, subagents and code, all of which count against the task's call budget.
Prime keeps its own runtime: a code agent that acts through a persistent interpreter and a shell inside a sandbox, may issue several calls per turn, and reaches the case system through a scripted client.
Its base agent is therefore stronger than the ReAct loop, and any CH-versus-Prime difference includes the runtime as well as the state layer.

\paragraph{When and how each harness learns.}
Prime opens a fresh session per ticket; after grading, the score line enters the same session and the agent takes a post-score turn in which it may read but not act, and writes its memory, skill and prompt files itself.
Mem0 can write memories during the task and takes a post-outcome memory turn after every serving ticket, with the score line in the same conversation; the stock Mem0 has no such turn, and its extraction model runs with Mem0's own defaults.
CH's acting agent never sees its score; the score reaches only the Refiner, which runs on the original's cadence (every 25 model steps for the first 200 steps, then every 100), evaluated between tickets, and revises the prompt, memories, skills and subagents in four model calls per pass.
RAG adds a block of twelve served cases, each with its score line, to a BM25 index once the block ends and retrieves the eight most similar cases for each new task.
SkillOpt treats every block of twelve serving tickets as one optimization step: the first six are the training batch and the last six the selection set on which a candidate skill is replayed for selection before it can replace the current skill.
The adapter maps each environment window to one optimization epoch: it runs a slow update after that window's serving tasks, samples replay tasks from the same window, and schedules meta-skill updates on the same epoch cycle (skipping the first epoch). This gives the optimizer access to window segmentation through scheduling and replay selection, even though model prompts contain no explicit window labels or hidden-policy descriptions.
To adapt the original fixed-validation setup to a stream, the selection set slides with the served tasks and the skill starts empty; each incoming task receives one scored rollout, and the strict acceptance gate accepts only about one step in nine.
Its validation replays re-run tasks that have already been served, about 0.85 additional task executions per served ticket, which is possible in a simulator but not in a deployment where each ticket is a live interaction, as in Pitch.

\paragraph{State and capacity.}
Prime injects a bounded summary of its harness files into the prompt and can search all of its past transcripts; CH shows overviews of its stores and bounds the Refiner's window; SkillOpt keeps one skill document with a per-step edit budget.
These capacities are not equal and are part of each method.

\paragraph{Accounting.}
Learning-side calls (Mem0's memory turn and extraction model, CH's Refiner passes, SkillOpt's optimizer and replays) are booked as maintenance.
Because Prime's relay ledger books its post-score turn together with the ticket turns, the learning share reported for Prime in Figure~\ref{fig:run-cost} is recovered from its per-ticket session records: every assistant message carries its own usage, and the runner's outcome message (\texttt{Ticket outcome.\ Score: $n$/100}) marks where the ticket work ends and the learning turn begins; the ticket-side totals so obtained equal the per-task usage in the result files exactly, and the ledger's remainder is counted as serving.
SkillOpt's validation replays use serving-side tasks only and never contribute to test scores.

\subsection{Evaluation Protocol}
\label{app:evaluation-protocol}

\paragraph{Evaluator and learner protocol.}
The evaluator has access to the full benchmark state, including the latent environment policy, task type, protocol role, and window boundaries. Learning harnesses receive task-visible information and accumulate serving interactions and outcome feedback; hidden-policy descriptions and corrections are withheld. Harness-specific use of window structure for update scheduling and replay is described in Appendix~\ref{app:learning-harnesses}. Learners may update their persistent state from serving experience, whereas test tasks reveal no feedback and do not affect subsequent adaptation. The objective is to improve performance on Hidden-Dependent tasks while preserving performance on Fully Specified tasks, thereby testing whether agents can learn hidden environment knowledge without disrupting behavior that is already correct.

\paragraph{Model-visible inputs and optimizer access.}
The acting and reflection components may receive: the visible task input, the agent's own tool trajectory or generated response, scalar outcome feedback, monotonic stream timestamp, and the learner's own persistent state.
They may not receive: slice labels, latent-state values, rule-family labels, oracle bulletins, corrected actions, evaluator explanations, or raw world-event time.
Window segmentation may be used by a harness's update scheduler or replay sampler, as specified in Appendix~\ref{app:learning-harnesses}; the active policy values and the content of policy changes remain hidden.
The benchmark's release gate audits model-visible prompts and metadata against these content restrictions.

\paragraph{Serving-side and test-side isolation.}
All tasks in a domain--tier instance belong to one evaluator-ordered EESD partitioned into environment windows.
Each window contains a serving side followed by a test side under the same latent environment state, and both sides cover every behavioral component exercised by the tier with disjoint concrete instances.
Each serving-side task is scored once with the candidate active when it arrives; its observed outcome may be used by the update rule. Subsequent replay scores, where supported, are used for harness optimization and are not included in reported serving or test scores.
After the serving tasks and any scheduled harness updates, the evaluator freezes the candidate state and runs every test-side probe independently from that state before the next environment state takes effect.
Probe-local state is discarded, and test-side trajectories and scores are excluded from every learner update, reflection minibatch, validation reservoir, and candidate-selection decision.
Only serving-side tasks receive learner-visible monotonic timestamps; the evaluator's test block does not create a visible timestamp gap.

\subsection{Data Processing and Reporting}
\label{app:data-reporting}
Table~\ref{tab:main-results} is generated from the benchmark's completed-cell results, snapshot dated September 25, 2026; the values are pinned and the table is rendered and checked by a script rather than typed.
Only current completed cells enter the table; archived superseded runs are excluded.
Every performance value in Table~\ref{tab:main-results} is a held-out test aggregate.
\paragraph{Aggregation convention.}
Cross-setting reward summaries first compute a mean task reward within each model--harness--setting cell and then average cells with equal weight. The tasks included depend on the reported phase and slice: Retail and Banking rewards are binary and scaled to 0--100, whereas Pitch uses the 0--100 rubric score.
Table~\ref{tab:main-results} uses held-out test tasks: Avg.~H averages the nine Hidden setting scores and Avg.~F the six Retail/Banking Fully Specified scores. Figure~\ref{fig:avg-performance} averages these test scores over the relevant models or harnesses, and Figure~\ref{fig:cost-frontier} directly uses the table's Avg.~H.
Figure~\ref{fig:cost-merged} pools serving and test tasks within each setting before averaging across settings; Figure~\ref{fig:cost-correctness} and Table~\ref{tab:cost-correctness} instead compute serving and test rewards separately. These displays share the table's cross-setting weighting, but only their Hidden test summaries use the same task subset as Avg.~H.
The reward axes in the exploration and knowledge-acquisition analyses (Section~\ref{sec:results-exploration}) use only the six Retail/Banking Hidden test settings, excluding Pitch; per-domain comparisons use the three settings of that domain. Their learning-dynamics metrics are computed from serving trajectories using the policy- and ticket-level definitions in Section~\ref{sec:analysis}, not by the cross-setting reward formula.
Item-weighted pooling would instead let the settings with the most test items dominate: Retail holds 1{,}686 of the 3{,}210 test items (53\%) and has the lowest Hidden scores, so item-weighted Hidden averages are several points lower for every harness (for example 30.7 rather than 38.3 for Mem0 and 60.2 rather than 61.7 for Prime over the four open-weight models). We therefore use equal setting weights for cross-setting reward summaries.
Averaging Retail and Banking accuracies with the Pitch rubric score treats each setting's score as a 0--100 reward and is a choice of the benchmark; the per-setting values remain available in Table~\ref{tab:main-results}.
Costs and token counts are not averaged this way: they are totals over all tasks of the lanes concerned, divided by the number of tasks, so a per-task cost is the mean over tasks.
Table~\ref{tab:main-results} computes its averages from the published one-decimal setting scores and rounds to one decimal; an incomplete set of constituent settings yields a dash.
The resource figures and Table~\ref{tab:cost-correctness} are computed from the archived result files of the completed lanes (bundle of September 25, 2026) with the same convention applied to the raw per-item rewards, which can move a value by 0.1 relative to the table.
Table~\ref{tab:cost-correctness} lists the per-task costs and reward values behind Figure~\ref{fig:cost-correctness}, aggregated by model and by learning harness over the same lanes per column (45 per model column, 36 per harness column).

\input{tex/cost_correctness_table}
Each cell summarizes one run; no standard error or confidence interval is implied.
Pitch L1 has only 6--8 effective attribute combinations per 24-item environment window, so the item count should not be interpreted as independent statistical replication.
\FloatBarrier
\normalsize

%% file: tex/cost_correctness_table.tex
% Generated table; edit the pinned source, not this file.
\begin{table}[t]
\centering
\footnotesize
\setlength{\tabcolsep}{4pt}
\caption{\textbf{Average cost per task and reward, by model and by learning harness.} Serving: API cost of one serving task including the harness's learning work (Refiner passes, memory model, optimizer, Prime's post-score turn), and reward on the serving stream. Test: API cost of one held-out test task under the frozen learned state, and test reward. Cost is total dollars over total tasks; reward is the mean over the column's model--harness--setting cells of the cell's Hidden (Adapt) or all-case (Total) reward, so every setting has equal weight as in Table~\ref{tab:main-results} (binary in Retail and Banking, the rubric rating in Pitch). Every model column aggregates 45 complete lanes (five harnesses $\times$ nine suites) and every harness column 36 (four models $\times$ nine suites); Blind and Oracle are excluded. Fireworks serverless prices of 2026-09-06.}
\label{tab:cost-correctness}
\begin{tabular}{l rrr rrr}
\toprule
 & \multicolumn{3}{c}{Serving (incl.\ learning)} & \multicolumn{3}{c}{Test (frozen state)} \\
\cmidrule(lr){2-4}\cmidrule(lr){5-7}
 & \$/task & Adapt & Total & \$/task & Adapt & Total \\
\midrule
\multicolumn{7}{l}{\textit{By model}} \\
GLM-5.3 Flash & 0.0117 & 40.4 & 59.5 & 0.0086 & 44.6 & 58.2 \\
DeepSeek V4.1 Flash & 0.0228 & 41.8 & 61.0 & 0.0170 & 44.9 & 60.1 \\
GLM-5.3 & 0.1732 & 48.5 & 64.5 & 0.1153 & 52.5 & 64.3 \\
Kimi K3 & 0.1880 & 48.9 & 63.9 & 0.1298 & 50.7 & 62.1 \\
\midrule
\multicolumn{7}{l}{\textit{By learning harness}} \\
SkillOpt & 0.0388 & 33.2 & 56.7 & 0.0180 & 39.8 & 55.9 \\
RAG & 0.0658 & 43.2 & 60.5 & 0.0832 & 43.4 & 57.4 \\
Mem0 & 0.0691 & 36.3 & 59.1 & 0.0213 & 38.3 & 56.0 \\
CH & 0.0891 & 54.6 & 66.2 & 0.0736 & 57.7 & 66.7 \\
Prime & 0.2320 & 57.2 & 68.6 & 0.1423 & 61.7 & 69.9 \\
\bottomrule
\end{tabular}
\end{table}

%% file: tex/a_benchmark_details.tex
\section{Detailed Benchmark Specification}
\label{app:benchmark-details}

This appendix specifies the learner-facing task interfaces, tier and window design, evaluator-side hidden-state schedules, and task-success verification.
The latent-state component descriptions below are evaluator metadata and are not provided to learning harnesses. Harness-specific use of window structure is described in Appendix~\ref{app:learning-harnesses}.

\subsection{Task and Scenario Interfaces}
\label{app:interfaces}

Table~\ref{tab:app-task-interfaces} separates the surface task from the latent object that changes.
Retail and banking require tool-mediated decisions whose correctness is verified from the final environment state; pitch requires a constrained natural-language response scored by a fixed rubric.

\begin{table}[htbp]
\centering
\scriptsize
\renewcommand{\arraystretch}{1.12}
\caption{Task interfaces and latent environment components.}
\label{tab:app-task-interfaces}
\begin{tabular}{@{}P{0.10\linewidth}P{0.28\linewidth}P{0.25\linewidth}P{0.28\linewidth}@{}}
\toprule
Domain & Task families & Required output & Latent state components \\
\midrule
Retail
& Cancel a pending order; exchange or return delivered items; modify a pending order; resolve multi-item and ranked-fallback requests.
& API actions that produce the correct database state, or a refusal with the correct reason code.
& Categorical eligibility scopes over payment form, product category, destination, refund method, order composition, and exception status. \\
Banking
& Authorize a pending card transaction; review a credit-limit increase; screen an outbound transfer; resolve ranked alternatives and payment batches in L2/L3.
& Approve, deny, or escalate with the correct reason; execute the first admissible alternative or screen each batch line when applicable.
& Numeric thresholds and scoped restrictions over merchant corridor, category limits, daily totals, account standing, requested limits, and payee age. \\
Pitch
& Write a 40--80 word sales pitch from a factual product sheet.
& A pitch that foregrounds positively valued attributes, suppresses negatively valued attributes, and introduces no unsupported claims.
& Valence over product attribute--value pairs; the active preference table is hidden and only the rubric score is returned. \\
\bottomrule
\end{tabular}
\end{table}
\FloatBarrier

\subsection{Tier Semantics and Window Sizes}
\label{app:tiers}

The tier labels describe temporal and compositional demands rather than a monotonic increase in surface-form complexity (Table~\ref{tab:app-tier-semantics}).
Within a tier, serving and test cover the same active behavioral components using disjoint instances.

\begin{table}[htbp]
\centering
\scriptsize
\renewcommand{\arraystretch}{1.12}
\caption{Cross-domain interpretation of the three difficulty tiers.}
\label{tab:app-tier-semantics}
\begin{tabular}{@{}P{0.06\linewidth}P{0.26\linewidth}P{0.31\linewidth}P{0.28\linewidth}@{}}
\toprule
Tier & Temporal demand & Within-task demand & Primary diagnostic \\
\midrule
L1
& Discover a latent state or newly introduced rule from outcome feedback; Pitch presents four profiles without re-entry.
& Mostly single-rule decisions; Pitch uses five visible attribute dimensions.
& Initial acquisition relative to the Blind--Oracle gap. \\
L2
& Revise beliefs after tightening, relaxation, scope migration, or limited profile re-entry.
& Scoped rules, correlated changes, ranked fallbacks, multi-line cases, or ten-dimensional preference selection.
& Hidden-state adaptation and impact on fully specified cases. \\
L3
& Track oscillation, rule removal, and re-entry of previously observed states.
& Multiple simultaneously relevant components and staggered reversals; retail also includes permissive departures from the document.
& Returned fully specified behavior, stale beliefs, and interference. \\
\bottomrule
\end{tabular}
\end{table}
\FloatBarrier

Table~\ref{tab:app-window-counts} gives the exact number of serving and test tasks in every released window.
Each row is one domain--tier scenario, \#W is its number of released windows, and an entry $S/T$ denotes $S$ serving tasks followed by $T$ test probes under the same environment state.

\newcommand{\nowindow}{\textcolor{gray}{n/a}}

\begin{table}[htbp]
\centering
\small
\setlength{\tabcolsep}{1.8pt}
\renewcommand{\arraystretch}{1.10}
\caption{Sample counts by scenario, difficulty, and environment window. Each entry is serving/test; the final row reports 53 windows and 4,508/3,210 tasks.}
\label{tab:app-window-counts}
\begin{tabular}{@{}llrccccccccr@{}}
\toprule
Domain & Tier & \#W & W0 & W1 & W2 & W3 & W4 & W5 & W6 & W7 & Total $S/T$ \\
\midrule
Retail & L1 & 6 & 79/44 & 140/64 & 140/64 & 120/64 & 120/64 & 121/84 & \nowindow & \nowindow & 720/384 \\
& L2 & 7 & 99/81 & 128/104 & 135/115 & 131/128 & 134/130 & 90/111 & 93/96 & \nowindow & 810/765 \\
& L3 & 8 & 78/60 & 75/65 & 84/68 & 78/67 & 77/76 & 74/72 & 80/64 & 78/65 & 624/537 \\
\midrule
Banking & L1 & 4 & 84/41 & 84/47 & 84/42 & 84/48 & \nowindow & \nowindow & \nowindow & \nowindow & 336/178 \\
& L2 & 5 & 100/74 & 100/77 & 100/76 & 100/80 & 100/80 & \nowindow & \nowindow & \nowindow & 500/387 \\
& L3 & 6 & 100/72 & 100/90 & 100/89 & 100/97 & 100/105 & 100/98 & \nowindow & \nowindow & 600/551 \\
\midrule
Pitch & L1 & 4 & 54/24 & 54/24 & 54/24 & 54/24 & \nowindow & \nowindow & \nowindow & \nowindow & 216/96 \\
& L2 & 6 & 54/24 & 54/24 & 54/24 & 54/24 & 54/24 & 54/24 & \nowindow & \nowindow & 324/144 \\
& L3 & 7 & 54/24 & 54/24 & 54/24 & 54/24 & 54/24 & 54/24 & 54/24 & \nowindow & 378/168 \\
\midrule
All & 9 tiers & 53 & \multicolumn{8}{c}{53 environment windows across all scenarios} & 4508/3210 \\
\bottomrule
\end{tabular}
\end{table}
\FloatBarrier

\newcommand{\policyon}{\ding{51}}
\newcommand{\policyoff}{\textcolor{gray}{\ding{55}}}

\subsection{Retail Hidden Rules and Activation Schedule}
\label{app:retail-rules}

The L1/L2/L3 scenarios use 7/14/12 categorical hidden rules over payment form, product category, destination, refund method, order composition, and exceptions.
We first define each atomic retail rule in Table~\ref{tab:app-retail-policy-defs}; Table~\ref{tab:app-retail-policy-active} then reports only whether that rule is active in each window. A block rule overrides a documented permission, whereas a permit rule introduces an exception to a documented restriction.

{\scriptsize
\setlength{\tabcolsep}{4pt}
\renewcommand{\arraystretch}{1.08}
\begin{longtable}{@{}llP{0.73\linewidth}@{}}
\caption{Dictionary of atomic retail hidden rules.}\label{tab:app-retail-policy-defs}\\
\toprule
Tier & ID & Meaning when active \\
\midrule
\endfirsthead
\toprule
Tier & ID & Meaning when active \\
\midrule
\endhead
L1 & H1 & Block modification of a pending order paid with multiple payment methods. \\
L1 & D1 & Block cancellation of a pending order paid with a gift card. \\
L1 & D2 & Block exchange of an electronics item. \\
L1 & D3 & Block exchange when the replacement destination is Florida. \\
L1 & D4 & Block refund to the original payment method. \\
L1 & D5 & Permit return of a clearance item. \\
L1 & D6 & Permit a return outside the documented return window. \\
\midrule
L2 & R1-E & Block exchange of an electronics item. \\
L2 & R1-A & Block exchange of an audio item. \\
L2 & R2-G & Block cancellation of a gift-card-paid order. \\
L2 & R2-S & Block cancellation of a split-paid order. \\
L2 & R3-All & Block every refund to the original card. \\
L2 & R3-Amex & Block refund to the original card only when that card is American Express. \\
L2 & R4-FL & Block exchange when the replacement destination is Florida. \\
L2 & R4-TX & Block exchange when the replacement destination is Texas. \\
L2 & R5 & Block modification of a split-paid order. \\
L2 & R6-M & Block a partial return from a multi-item order. \\
L2 & R6-ME & Block a partial return only when the multi-item order contains electronics. \\
L2 & R7 & Block a partial exchange. \\
L2 & R8 & Block exchange to a PO Box. \\
L2 & R9 & Permit a gift-card-funded exchange upcharge. \\
\midrule
L3 & Z1-A & Block exchange of an audio item. \\
L3 & Z1-E & Block exchange of an electronics item. \\
L3 & Z2-G & Block cancellation of a gift-card-paid order. \\
L3 & Z2-S & Block cancellation of a split-paid order. \\
L3 & Z3-O & Permit refund to the original payment method. \\
L3 & Z3-V & Permit refund to the original card only when that card is Visa. \\
L3 & Z4 & Permit return of a clearance item. \\
L3 & Z5 & Block modification of a split-paid order. \\
L3 & Z6 & Block exchange to a PO Box. \\
L3 & Z7-FL & Block exchange when the replacement destination is Florida. \\
L3 & Z7-TX & Block exchange when the replacement destination is Texas. \\
L3 & Z8 & Permit a gift-card-funded exchange upcharge. \\
\bottomrule
\end{longtable}
}

In the activation matrices, \policyon{} means active, \policyoff{} means inactive, and \nowindow{} means that the tier has no such window.

\begin{table}[htbp]
\centering
\scriptsize
\setlength{\tabcolsep}{3.7pt}
\renewcommand{\arraystretch}{1.06}
\caption{Retail hidden-rule activation by window. A check mark denotes active, a cross denotes inactive, and n/a denotes an unavailable window. Rule meanings are defined in Table~\ref{tab:app-retail-policy-defs}.}
\label{tab:app-retail-policy-active}
\begin{tabular}{@{}llcccccccc@{}}
\toprule
Tier & ID & W0 & W1 & W2 & W3 & W4 & W5 & W6 & W7 \\
\midrule
L1 & H1 & \policyon & \policyon & \policyon & \policyon & \policyon & \policyon & \nowindow & \nowindow \\
L1 & D1 & \policyoff & \policyon & \policyon & \policyoff & \policyoff & \policyoff & \nowindow & \nowindow \\
L1 & D2 & \policyoff & \policyoff & \policyon & \policyon & \policyon & \policyon & \nowindow & \nowindow \\
L1 & D3 & \policyoff & \policyoff & \policyoff & \policyoff & \policyon & \policyon & \nowindow & \nowindow \\
L1 & D4 & \policyoff & \policyoff & \policyoff & \policyoff & \policyoff & \policyon & \nowindow & \nowindow \\
L1 & D5 & \policyon & \policyon & \policyon & \policyon & \policyon & \policyon & \nowindow & \nowindow \\
L1 & D6 & \policyoff & \policyoff & \policyoff & \policyon & \policyon & \policyon & \nowindow & \nowindow \\
\midrule
L2 & R1-E & \policyoff & \policyon & \policyon & \policyoff & \policyoff & \policyoff & \policyoff & \nowindow \\
L2 & R1-A & \policyoff & \policyoff & \policyoff & \policyon & \policyon & \policyoff & \policyoff & \nowindow \\
L2 & R2-G & \policyoff & \policyoff & \policyon & \policyon & \policyoff & \policyoff & \policyoff & \nowindow \\
L2 & R2-S & \policyoff & \policyoff & \policyoff & \policyoff & \policyon & \policyon & \policyoff & \nowindow \\
L2 & R3-All & \policyoff & \policyoff & \policyon & \policyon & \policyoff & \policyoff & \policyoff & \nowindow \\
L2 & R3-Amex & \policyoff & \policyoff & \policyoff & \policyoff & \policyon & \policyon & \policyon & \nowindow \\
L2 & R4-FL & \policyoff & \policyoff & \policyoff & \policyon & \policyon & \policyoff & \policyoff & \nowindow \\
L2 & R4-TX & \policyoff & \policyoff & \policyoff & \policyoff & \policyoff & \policyon & \policyon & \nowindow \\
L2 & R5 & \policyon & \policyon & \policyon & \policyon & \policyon & \policyon & \policyon & \nowindow \\
L2 & R6-M & \policyoff & \policyon & \policyon & \policyon & \policyoff & \policyoff & \policyoff & \nowindow \\
L2 & R6-ME & \policyoff & \policyoff & \policyoff & \policyoff & \policyon & \policyon & \policyon & \nowindow \\
L2 & R7 & \policyoff & \policyoff & \policyon & \policyon & \policyon & \policyon & \policyoff & \nowindow \\
L2 & R8 & \policyon & \policyon & \policyon & \policyon & \policyon & \policyon & \policyon & \nowindow \\
L2 & R9 & \policyon & \policyon & \policyon & \policyon & \policyon & \policyon & \policyon & \nowindow \\
\midrule
L3 & Z1-A & \policyoff & \policyon & \policyoff & \policyoff & \policyoff & \policyoff & \policyoff & \policyoff \\
L3 & Z1-E & \policyoff & \policyoff & \policyoff & \policyon & \policyon & \policyoff & \policyoff & \policyoff \\
L3 & Z2-G & \policyoff & \policyon & \policyoff & \policyon & \policyon & \policyoff & \policyoff & \policyoff \\
L3 & Z2-S & \policyoff & \policyoff & \policyon & \policyoff & \policyoff & \policyon & \policyon & \policyoff \\
L3 & Z3-O & \policyoff & \policyoff & \policyon & \policyoff & \policyoff & \policyoff & \policyoff & \policyoff \\
L3 & Z3-V & \policyoff & \policyoff & \policyoff & \policyoff & \policyon & \policyon & \policyoff & \policyoff \\
L3 & Z4 & \policyoff & \policyoff & \policyon & \policyoff & \policyoff & \policyoff & \policyon & \policyon \\
L3 & Z5 & \policyon & \policyon & \policyon & \policyon & \policyon & \policyon & \policyon & \policyon \\
L3 & Z6 & \policyon & \policyon & \policyon & \policyon & \policyon & \policyon & \policyon & \policyon \\
L3 & Z7-FL & \policyoff & \policyon & \policyoff & \policyoff & \policyoff & \policyon & \policyon & \policyoff \\
L3 & Z7-TX & \policyoff & \policyoff & \policyoff & \policyon & \policyoff & \policyoff & \policyoff & \policyoff \\
L3 & Z8 & \policyon & \policyon & \policyon & \policyon & \policyon & \policyon & \policyon & \policyon \\
\bottomrule
\end{tabular}
\end{table}
\clearpage

\subsection{Banking Hidden Rules and Activation Schedule}
\label{app:banking-rules}

The L1/L2/L3 scenarios use 5/9/11 rule-value states across 4/6/6 policy dimensions; verification checks the submitted verdict and terminal environment state.
Table~\ref{tab:app-banking-policy-defs} turns each threshold or categorical scope into an atomic rule. Table~\ref{tab:app-banking-policy-active} then marks the windows in which that exact value is used.

{\scriptsize
\setlength{\tabcolsep}{4pt}
\renewcommand{\arraystretch}{1.08}
\begin{longtable}{@{}llP{0.73\linewidth}@{}}
\caption{Dictionary of atomic banking hidden rules.}\label{tab:app-banking-policy-defs}\\
\toprule
Tier & ID & Meaning when active \\
\midrule
\endfirsthead
\toprule
Tier & ID & Meaning when active \\
\midrule
\endhead
L1 & S1 & Basic verification is insufficient for a gift-card purchase. \\
L1 & S2-650 & Set the electronics instant-approval ceiling to \$650. \\
L1 & D1-J & Apply the overseas-merchant restriction to jewelry. \\
L1 & D2-3 & Use the three-payee restricted-payee roster. \\
L1 & D2-5 & Use the five-payee restricted-payee roster. \\
\midrule
L2 & S1-2000 & Set the same-day transaction-total cap to \$2,000. \\
L2 & S2-6500 & Set the review threshold for recently added payees to \$6,500. \\
L2 & D1-650 & Set the electronics instant-approval ceiling to \$650. \\
L2 & D1-460 & Set the electronics instant-approval ceiling to \$460. \\
L2 & D2-J & Apply the overseas-merchant restriction to jewelry. \\
L2 & D2-E & Apply the overseas-merchant restriction to electronics. \\
L2 & D3-900 & Set the travel instant-approval ceiling to \$900. \\
L2 & D3-1150 & Set the travel instant-approval ceiling to \$1,150. \\
L2 & D4-3800 & Permit credit-limit increases up to \$3,800. \\
\midrule
L3 & P1-650 & Set the electronics instant-approval ceiling to \$650. \\
L3 & P1-460 & Set the electronics instant-approval ceiling to \$460. \\
L3 & P2-900 & Set the travel instant-approval ceiling to \$900. \\
L3 & P2-1150 & Set the travel instant-approval ceiling to \$1,150. \\
L3 & P3-J & Apply the overseas-merchant restriction to jewelry. \\
L3 & P3-E & Apply the overseas-merchant restriction to electronics. \\
L3 & P4-2000 & Set the same-day transaction-total cap to \$2,000. \\
L3 & P4-1700 & Set the same-day transaction-total cap to \$1,700. \\
L3 & P5-3800 & Permit credit-limit increases up to \$3,800. \\
L3 & P6-6500 & Set the review threshold for recently added payees to \$6,500. \\
L3 & P6-7200 & Set the review threshold for recently added payees to \$7,200. \\
\bottomrule
\end{longtable}
}

\begin{table}[htbp]
\centering
\scriptsize
\setlength{\tabcolsep}{5pt}
\renewcommand{\arraystretch}{1.06}
\caption{Banking hidden-rule activation by window. A check mark denotes active, a cross denotes inactive, and n/a denotes an unavailable window. Rule meanings are defined in Table~\ref{tab:app-banking-policy-defs}.}
\label{tab:app-banking-policy-active}
\begin{tabular}{@{}llcccccc@{}}
\toprule
Tier & ID & W0 & W1 & W2 & W3 & W4 & W5 \\
\midrule
L1 & S1 & \policyon & \policyon & \policyon & \policyon & \nowindow & \nowindow \\
L1 & S2-650 & \policyon & \policyon & \policyon & \policyon & \nowindow & \nowindow \\
L1 & D1-J & \policyoff & \policyon & \policyon & \policyon & \nowindow & \nowindow \\
L1 & D2-3 & \policyon & \policyon & \policyon & \policyoff & \nowindow & \nowindow \\
L1 & D2-5 & \policyoff & \policyoff & \policyoff & \policyon & \nowindow & \nowindow \\
\midrule
L2 & S1-2000 & \policyon & \policyon & \policyon & \policyon & \policyon & \nowindow \\
L2 & S2-6500 & \policyon & \policyon & \policyon & \policyon & \policyon & \nowindow \\
L2 & D1-650 & \policyon & \policyoff & \policyoff & \policyoff & \policyoff & \nowindow \\
L2 & D1-460 & \policyoff & \policyon & \policyon & \policyon & \policyon & \nowindow \\
L2 & D2-J & \policyon & \policyon & \policyoff & \policyoff & \policyoff & \nowindow \\
L2 & D2-E & \policyoff & \policyoff & \policyon & \policyon & \policyon & \nowindow \\
L2 & D3-900 & \policyon & \policyon & \policyon & \policyoff & \policyoff & \nowindow \\
L2 & D3-1150 & \policyoff & \policyoff & \policyoff & \policyon & \policyon & \nowindow \\
L2 & D4-3800 & \policyoff & \policyoff & \policyoff & \policyoff & \policyon & \nowindow \\
\midrule
L3 & P1-650 & \policyon & \policyoff & \policyoff & \policyon & \policyon & \policyon \\
L3 & P1-460 & \policyoff & \policyon & \policyon & \policyoff & \policyoff & \policyoff \\
L3 & P2-900 & \policyon & \policyon & \policyon & \policyoff & \policyoff & \policyon \\
L3 & P2-1150 & \policyoff & \policyoff & \policyoff & \policyon & \policyon & \policyoff \\
L3 & P3-J & \policyon & \policyon & \policyoff & \policyoff & \policyon & \policyoff \\
L3 & P3-E & \policyoff & \policyoff & \policyon & \policyon & \policyoff & \policyoff \\
L3 & P4-2000 & \policyon & \policyon & \policyon & \policyon & \policyoff & \policyoff \\
L3 & P4-1700 & \policyoff & \policyoff & \policyoff & \policyoff & \policyon & \policyon \\
L3 & P5-3800 & \policyoff & \policyon & \policyon & \policyon & \policyoff & \policyoff \\
L3 & P6-6500 & \policyon & \policyon & \policyoff & \policyoff & \policyoff & \policyoff \\
L3 & P6-7200 & \policyoff & \policyoff & \policyon & \policyon & \policyon & \policyon \\
\bottomrule
\end{tabular}
\end{table}
\clearpage

\subsection{Pitch Latent Preferences and Activation Schedule}
\label{app:pitch-rules}

Each Pitch window activates exactly one complete customer preference profile. Table~\ref{tab:app-pitch-policy-defs} defines the six profiles used across the benchmark, and Table~\ref{tab:app-pitch-policy-active} reports their use by tier and window.
Every window contains 54 serving and 24 test questions (Table~\ref{tab:app-window-counts}).
L1 uses four non-repeating profiles over four windows and five visible attributes; L2 uses four profiles over six windows, revisits two profiles, and expands to ten attributes; L3 uses four profiles over seven windows with repeated returns and ten attributes.

\begin{table}[htbp]
\centering
\scriptsize
\renewcommand{\arraystretch}{1.10}
\caption{Dictionary of latent pitch preferences.}
\label{tab:app-pitch-policy-defs}
\begin{tabular}{@{}lP{0.78\linewidth}@{}}
\toprule
ID & Meaning when active \\
\midrule
P-Val & Use the value-for-money buyer profile. \\
P-Id & Use the identity-expressive buyer profile. \\
P-AM & Use the anti-marketing rationalist profile. \\
P-RA & Use the risk-averse buyer profile. \\
P-Eco & Use the eco-conscious buyer profile. \\
P-Conf & Use the conformist buyer profile. \\
\bottomrule
\end{tabular}
\end{table}

\begin{table}[htbp]
\centering
\scriptsize
\setlength{\tabcolsep}{4.3pt}
\renewcommand{\arraystretch}{1.08}
\caption{Pitch preference activation by window. A check mark denotes active, a cross denotes inactive, and n/a denotes an unavailable window. Preference meanings are defined in Table~\ref{tab:app-pitch-policy-defs}.}
\label{tab:app-pitch-policy-active}
\begin{tabular}{@{}llccccccc@{}}
\toprule
Tier & ID & W0 & W1 & W2 & W3 & W4 & W5 & W6 \\
\midrule
L1 & P-Val & \policyon & \policyoff & \policyoff & \policyoff & \nowindow & \nowindow & \nowindow \\
L1 & P-Id & \policyoff & \policyon & \policyoff & \policyoff & \nowindow & \nowindow & \nowindow \\
L1 & P-AM & \policyoff & \policyoff & \policyon & \policyoff & \nowindow & \nowindow & \nowindow \\
L1 & P-RA & \policyoff & \policyoff & \policyoff & \policyon & \nowindow & \nowindow & \nowindow \\
\midrule
L2 & P-RA & \policyon & \policyoff & \policyoff & \policyoff & \policyoff & \policyon & \nowindow \\
L2 & P-Val & \policyoff & \policyon & \policyoff & \policyoff & \policyon & \policyoff & \nowindow \\
L2 & P-Eco & \policyoff & \policyoff & \policyon & \policyoff & \policyoff & \policyoff & \nowindow \\
L2 & P-Id & \policyoff & \policyoff & \policyoff & \policyon & \policyoff & \policyoff & \nowindow \\
\midrule
L3 & P-RA & \policyon & \policyoff & \policyoff & \policyoff & \policyon & \policyoff & \policyoff \\
L3 & P-Id & \policyoff & \policyon & \policyoff & \policyon & \policyoff & \policyoff & \policyon \\
L3 & P-AM & \policyoff & \policyoff & \policyon & \policyoff & \policyoff & \policyoff & \policyoff \\
L3 & P-Conf & \policyoff & \policyoff & \policyoff & \policyoff & \policyoff & \policyon & \policyoff \\
\bottomrule
\end{tabular}
\end{table}
\FloatBarrier

Together, the dictionaries and binary matrices separate profile meaning from temporal activation, making discovery, revision, and re-entry directly visible.

\input{tex/a_task_success_verification}

%% file: tex/a_task_success_verification.tex
\subsection{Task Success Verification}
\label{app:task-evaluation}

\paragraph{Retail and Banking: deterministic verification.}
Retail and Banking share a rule-based, outcome-only verifier with no LLM judgment.
For task $n$, let $D_n^{\mathrm{agent}}$ be the terminal database produced by the agent.
The evaluator constructs the reference terminal database $D_n^\star$ by reloading a clean database, applying the same task-specific view, and replaying the ground-truth tool actions.
After canonicalizing dictionaries and sets as order-independent and lists as order-sensitive, it computes
\begin{equation}
  r_n^{\mathrm{state}}
  =\mathbb{I}\!\left[H(D_n^{\mathrm{agent}})=H(D_n^\star)\right],
  \label{eq:rule-state-check}
\end{equation}
where $H(\cdot)$ is a SHA-256 hash of the canonicalized database and $\mathbb{I}[\cdot]$ is the indicator function.
The verifier therefore scores the terminal outcome rather than requiring the agent to reproduce a particular sequence of intermediate actions.

The answer check $r_n^{\mathrm{answer}}$ requires exactly one structured \texttt{finish} action.
For state-changing tasks, the finish declaration must be consistent with carrying out the request; for refusal tasks, it must declare a refusal and provide one of the accepted reason codes; and for value tasks, it must match the expected typed value.
Integer answers must contain exactly one integral value, floating-point answers must contain exactly one value within $0.01$ of the reference, normalized strings must match exactly after lowercasing and removing non-alphanumeric characters, and sets are compared after element-wise normalization.
The binary task reward is
\begin{equation}
  R_n^{\mathrm{tool}}
  =r_n^{\mathrm{state}}r_n^{\mathrm{answer}}
  \in\{0,1\}.
  \label{eq:rule-task-reward}
\end{equation}
Thus a correct final answer cannot compensate for an incorrect database transition, and a correct database state cannot compensate for a missing or inconsistent finish declaration.

\paragraph{Pitch: structured LLM judgment.}
Pitch has no mutable database, so each submitted pitch is evaluated by a fixed GLM-5.3 Flash judge with high reasoning effort and provider-default sampling.
For every attribute on the product sheet, the judge returns one structured verdict: \textsc{M} if the pitch mentions or paraphrases the true attribute consistently, \textsc{C} if it directly contradicts or reverses the attribute, and \textsc{N} if it does not address the attribute.
The judge separately lists concrete factual claims unsupported by any product-sheet attribute.
Its output must be a complete JSON object covering every attribute; malformed or incomplete outputs fail closed and are retried rather than being interpreted as favorable judgments.
The LLM produces only this checklist, while deterministic code maps the active customer profile and checklist to the final numerical rating.

Let $\mathcal{A}_n$ be the product attributes for task $n$, $c_i\in\{\mathrm{strong},\mathrm{pos},\mathrm{neu},\mathrm{neg},\mathrm{red}\}$ the preference class of attribute $i$ under the active profile, and $v_i\in\{\mathrm{M},\mathrm{C},\mathrm{N}\}$ the judge verdict.
Strongly positive and positive attributes receive weights $w(\mathrm{strong})=2$ and $w(\mathrm{pos})=1$, with all other classes receiving zero positive-coverage weight.
The weighted positive coverage is
\begin{equation}
  C_n=
  \frac{\sum_{i\in\mathcal{A}_n}w(c_i)\,\mathbb{I}[v_i=\mathrm{M}]}
       {\sum_{i\in\mathcal{A}_n}w(c_i)},
  \label{eq:pitch-coverage}
\end{equation}
with $C_n=0$ if the denominator is zero.
Let $N_{\mathrm{neg}}$ and $N_{\mathrm{red}}$ count negative and red-line attributes, respectively, whose verdict is either \textsc{M} or \textsc{C}; let $N_{\mathrm{con}}$ count all \textsc{C} verdicts; and let $N_{\mathrm{fab}}$ be the number of distinct unsupported claims, capped at five.
The pre-length score is
\begin{equation}
  S_n^{\mathrm{raw}}
  =100C_n\,(0.5)^{N_{\mathrm{neg}}}(0.3)^{N_{\mathrm{red}}}
          (0.5)^{N_{\mathrm{con}}}(0.5)^{N_{\mathrm{fab}}}.
  \label{eq:pitch-raw-score}
\end{equation}
Although the requested pitch length is 40--80 words, the verifier allows a grace margin through 90 words.
For word count $L_n$, the final rating and normalized reward are
\begin{equation}
  S_n=
  \begin{cases}
    S_n^{\mathrm{raw}}, & L_n\le 90,\\
    \min(S_n^{\mathrm{raw}},25), & L_n>90,
  \end{cases}
  \qquad
  R_n^{\mathrm{pitch}}=\frac{S_n}{100}.
  \label{eq:pitch-final-score}
\end{equation}
The experiments report the continuous rating $S_n$; the implementation additionally records $S_n\ge75$ as the binary task-success flag.

%% file: tex/a_additional_evaluation.tex
\section{Additional Evaluation}
\label{app:additional-evaluation}
\small

This appendix details policy-level learning speed, the behavior of agents when a previously learned policy is withdrawn, answer diversity per domain, the breakdown of serving-stream costs, preliminary attempts at training-based learning, and interpretation of Fully Specified reference performance.

\subsection{Learning-Speed Analysis}
\label{app:learning-speed}

\paragraph{Policies.}
A policy is one rule of a Banking or Retail tier in one state, from the window in which that state takes effect (or window W0 for a rule present from the start) to the window before the rule next changes.
Intervals whose state agrees with the manual are skipped for Retail, since no Hidden ticket can arise from them.
Across the three tiers this yields 31 Banking and 37 Retail policies: 23 present from the start, 23 tightened, 9 relaxed, 11 re-scoped, one newly introduced, and one removed.

\paragraph{Attributing tickets to rules.}
A serving ticket belongs to a rule when that rule decides its correct outcome.
For Banking we use the evaluator-side, state-parameterised decision engine of the benchmark's learning-dynamics tools, which is checked to reproduce every locked verdict before use: a rule decides a ticket when setting that rule alone to another value changes the correct decision.
For Retail L1 each rule has exactly one Hidden task family, so the family identifies the rule.
For Retail L2 and L3 we apply the same counterfactual with each tier's own ground-truth engine: rule $r$ decides a ticket when replacing $r$'s true state by the state the manual describes, holding the other rules fixed, changes the correct actions or answer.
The attribution reproduces the recorded Hidden/Fully Specified label of every one of the 810 Retail L2 and 624 Retail L3 serving tickets.
Three Retail L2 Hidden tickets change outcome only when several rules change together and are left out.
The 27 Banking L3 serving tickets that the bundle files mark \emph{reopen} are classified by the same engine: 8 are Hidden and 19 Fully Specified.
Retail windows are taken from the rule timeline at the ticket's time, because the Retail L3 result rows number serving windows differently.

\paragraph{Tickets decided by several rules.}
Almost half of the Hidden serving tickets in Banking L2 and L3 are decided by two rules at once (82 of 172 and 101 of 198), mostly a daily spending cap together with a category limit, or the new-payee and reporting thresholds together; Retail has 9 such tickets.
Every attributed ticket counts for each rule that decides it, both as an example the harness has seen (the $k$ axis) and as a measurement ($a_k$).
Counting them as seen but measuring only single-rule tickets gives the same ranking of harnesses, but in Banking the set of policies behind each $a_k$ then changes with $k$, because some policies contribute single-rule tickets only late in their interval; this produces spurious dips in the curve.

\paragraph{Curves and AULC.}
$a_k$ pools all attributed tickets of a harness with exactly $k$ earlier tickets of the same policy, over policies, models, and tiers of a domain; panels (a, b) of Figure~\ref{fig:learning-speed} show positions with at least 20 tickets.
AULC$_{10}$ (Section~\ref{sec:results-learning-speed}) is computed per heat-map cell from the same tickets; a position with fewer than three tickets is dropped, and the cell is left empty if fewer than eight of the ten positions remain.
The scripts and the per-ticket data will be released with the benchmark.

\subsection{Learning a Policy versus Dropping It}
\label{app:learn-vs-drop}

A policy that has been learned becomes a liability once the rule returns to what the manual says: the harness must stop applying it.
We compare this with learning the same policy, in the same lane.
For every rule state $S$ that is followed by a change of the rule (31 states: 10 in Banking L2--L3, 21 in Retail L1--L3; 15 of them return the rule to the manual, the others re-scope it), the \emph{learning} side is the sequence of attributed Hidden serving tickets while $S$ is in force (Appendix~\ref{app:learning-speed}), and the \emph{dropping} side is the sequence of serving tickets after $S$ ends whose correct answer is the manual's and would be wrong under $S$.
Dropping tickets are therefore Fully Specified: nothing new has to be learned, only $S$ has to be given up.
They are found with the same counterfactual engines: Banking via the discriminating tickets of each rule change, Retail L2--L3 by re-deciding each general ticket with the rule set back to its previous state, and Retail L1 by the one lifted ban (gift-card cancellation, lifted at $t{=}360$).
Both sides come from one (model, harness, suite, rule, $S$) pair, so the policy, the model and the harness are identical.
We keep the pairs in which the harness had learned $S$ (at least 60\% correct on its last five tickets of $S$); without this condition, a harness that never learned $S$ has nothing to drop and scores highly on the dropping side for that reason alone.

\begin{figure}[t]
  \centering
  \includegraphics[width=\textwidth]{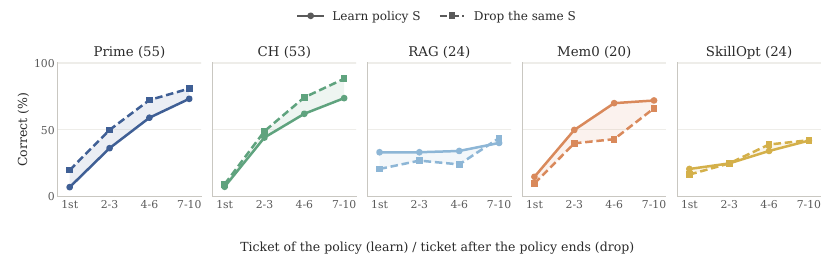}
  \caption{\textbf{Learning a policy versus dropping the same policy.}
  Solid: correctness on the $k$-th Hidden ticket of policy $S$. Dashed:
  correctness on the $k$-th ticket after $S$ ends that $S$ would answer
  wrongly. Same model, harness and policy on both sides; pairs in which $S$
  was learned; number of pairs in parentheses; four models, Banking and
  Retail pooled.}
  \label{fig:learn-vs-drop}
\end{figure}

Figure~\ref{fig:learn-vs-drop} shows that, for the same policy, dropping it is about as fast as learning it.
Prime starts higher when dropping (20\% on the first ticket, against 7\% when learning), CH barely (9\% against 8\%), and both end 8--14 points higher after seven to ten tickets (81\% and 88\%, against 73\% and 74\%); the two curves have the same shape.
On the first ticket after the change most harnesses still act on the superseded policy, although a harness that had learned nothing would follow the manual and answer these tickets correctly.
Mem0 is the exception: it learns $S$ faster than it drops it (70\% against 43\% on the fourth to sixth ticket).
RAG, whose exemplar store is also append-only, drops policies slowly as well.
This explanation is consistent with the curves but not verified at the level of retrieved memories, and the Mem0, RAG and SkillOpt panels rest on 20--24 pairs.

\subsection{Exploration by Domain}
\label{app:exploration}

\begin{figure}[t]
  \centering
  \includegraphics[width=\textwidth]{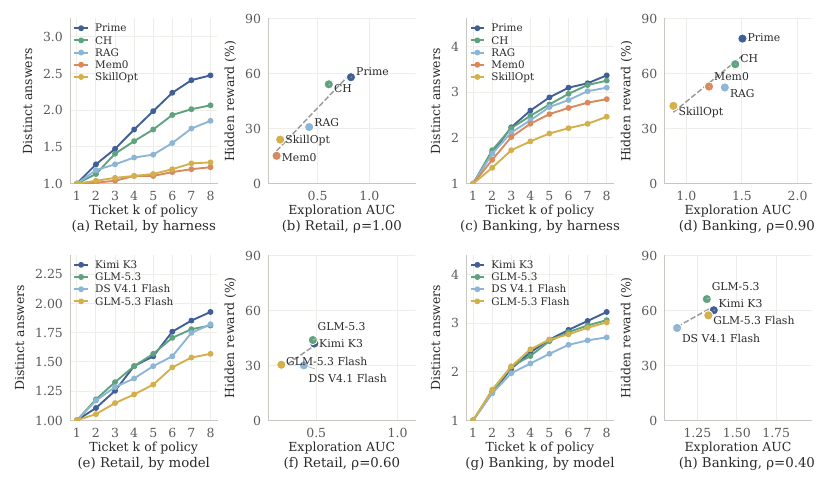}
  \caption{\textbf{Answer diversity and reward per domain.}
  \textbf{(a, c, e, g)} Mean number of distinct answers given on a policy after $k$ of its Hidden serving tickets, by harness (top) and by model (bottom); correctness is ignored.
  \textbf{(b, d, f, h)} Exploration AUC of those curves against the domain's average reward on Hidden cases, with the Spearman $\rho$ and a least-squares fit.}
  \label{fig:exploration-by-domain}
\end{figure}

Figure~\ref{fig:exploration-by-domain} shows the answer-diversity curves and the reward relation of Figure~\ref{fig:exploration} for Retail and Banking separately.
The sequences are the attributed Hidden serving tickets of each policy (Appendix~\ref{app:learning-speed}); for Retail we use the restrictive policies, whose correct answer is a refusal, and for Banking the tickets decided by a single rule.
A Retail answer is \emph{execute} or \emph{refuse} with one of the accepted refusal codes; the few Retail tickets whose correct answer is a value rather than an action are excluded; a Banking answer is one of approve, deny or escalate together with its reason code, so the answer space is larger in Banking and its curves are higher.
Only policies with at least $K=8$ tickets enter a curve (Section~\ref{sec:results-exploration}), so every curve averages the same policies at every $k$: 20 of 20, 36 of 52 and 20 of 44 policies per harness in Retail L1--L3 and 20 of 20, 12 of 20 and 20 of 24 in Banking L1--L3; the excluded policies are the short ones of Retail L2--L3 and Banking L2--L3, and policies longer than $K$ (median 42 tickets in Retail L1, up to 109) contribute only their first $K$.
The horizon is a trade-off between reading more of the curve and keeping more policies: for every $K$ from 4 to 12 (112 to 44 Retail and 64 to 40 Banking policies per harness) the harness order of the exploration AUC is Prime, CH, RAG, Mem0, SkillOpt, its rank correlation with average Hidden reward is 1.00, and the model-level correlation is 0.80 (1.00 at $K=6$--7).
SkillOpt revises its skill once per block of twelve serving tickets, so within a block its answers can change only through the model's own variation; its curves are therefore a lower bound on the diversity a per-ticket update would show.

\begin{figure}[t]
  \centering
  \includegraphics[width=\textwidth]{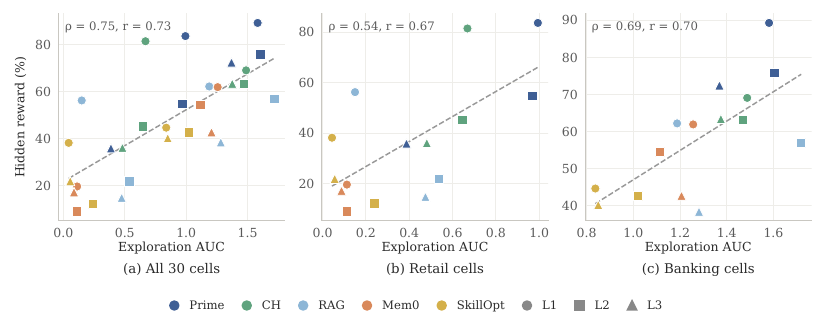}
  \caption{\textbf{Exploration and reward per harness, domain and tier.}
  One point per (harness, domain, tier) cell: exploration AUC (Section~\ref{sec:results-exploration}, $K=8$) over the cell's policies with the four open-weight models pooled, against the cell's reward on Hidden cases from Table~\ref{tab:main-results} averaged over the same models.
  \textbf{(a)} all 30 cells, \textbf{(b)} Retail, \textbf{(c)} Banking; colour is the harness, shape the tier; $\rho$ and $r$ are the Spearman and Pearson correlations, the dashed line a least-squares fit.}
  \label{fig:exploration-cells}
\end{figure}

The harness ranking of Figure~\ref{fig:exploration}(a) holds in both domains except that Mem0 and SkillOpt swap in Retail.
In Retail the exploration AUCs are 0.82 (Prime), 0.61 (CH), 0.42 (RAG), 0.14 (SkillOpt) and 0.11 (Mem0), against average Hidden rewards of 58.1, 54.2, 30.9, 24.0 and 15.2 ($\rho = 1.00$); in Banking they are 1.50, 1.44, 1.35, 1.20 (Mem0) and 0.88 (SkillOpt), against 79.2, 65.2, 52.5, 53.0 and 42.5 ($\rho = 0.90$).
By model the correlations are 0.60 in Retail and 0.40 in Banking, over four points.
Figure~\ref{fig:exploration-cells} repeats the comparison one level finer, with one point per harness, domain and tier (30 cells): the rank correlation is 0.75 over all cells, 0.54 within Retail and 0.69 within Banking, and 0.72, 0.90 and 0.87 within L1, L2 and L3.
The finer split adds variance that has nothing to do with the harness, since tiers differ in reward by up to 60 points and the two domains have answer spaces of different size, so the correlation is lower than at the harness level while the direction is the same in every subset; the Retail outliers are RAG, which explores little on L1 yet scores 56\% and explores more on L3 yet scores 15\%, consistent with its retention problem in Figure~\ref{fig:consolidation}.

At the finer level of the 20 model--harness cells of a domain, the AUC correlates with the cell's Hidden reward at $\rho = 0.92$ in Retail and 0.57 in Banking; the Banking value is lower because fast-learning cells stop varying their answers early.
A variant that counts, over the tickets strictly before a policy's first correct one (policies with at least two such tickets), the number of distinct answers divided by the number of tickets, pooled over a cell's policies, reaches $\rho = 0.94$ and 0.73, and keeps a median within-model correlation of 0.90 and 0.75 across harnesses, so the relation is not explained by stronger models both exploring more and scoring higher.
The knowledge-acquisition curves of Figure~\ref{fig:consolidation} average Retail and Banking with equal weight; per harness they rest on 49/41 (Prime), 49/40 (CH), 40/40 (RAG), 6/33 (Mem0) and 13/30 (SkillOpt) Retail/Banking policies, so the Mem0 value is the least certain.
The scripts and per-policy sequences will be released with the benchmark.

% \FloatBarrier
\subsection{Cost per Phase and Learning Cost Breakdown}
\label{app:cost-breakdown}

\begin{figure*}[t]
  \centering
  \includegraphics[width=\textwidth]{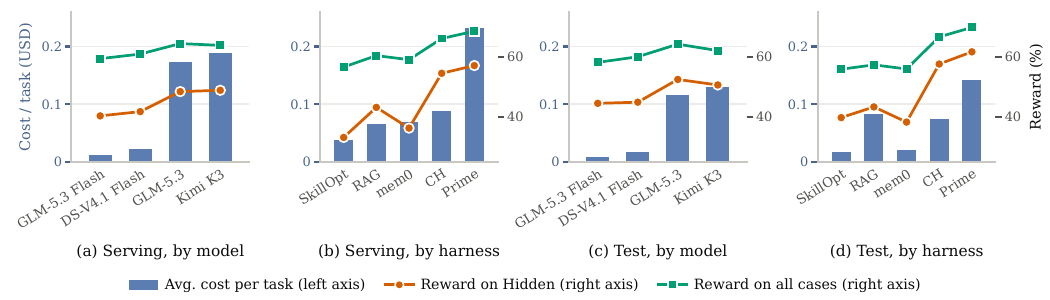}
  \caption{\textbf{Average cost per task against reward, by phase.}
  Bars (left axis) give the API cost of one task; lines (right axis) give
  reward on Hidden (adapt) cases and on all cases, averaged over the nine suites with equal weight.
  \textbf{(a, b)} Serving stream: cost includes the harness's learning work,
  reward is measured on the serving cases. \textbf{(c, d)} Held-out test
  under the frozen learned state. Columns are grouped by model (a, c) and
  by learning harness (b, d), ordered by serving cost; every column aggregates
  the same lanes (45 per model column, 36 per harness column). Blind and Oracle are excluded.}
  \label{fig:cost-correctness}
\end{figure*}

Figure~\ref{fig:cost-correctness} shows the comparison of Figure~\ref{fig:cost-merged} with the serving stream and the held-out test kept apart.
Serving cost charges everything the harness spends while the stream runs, the model's own turns and the learning work, divided by the number of serving tasks, and is paired with reward on the serving cases; test cost is the inference cost of one held-out task under the frozen learned state, paired with test reward.
Reward is averaged equally over the column's model--harness--setting cells, separately for each phase. Hidden test values match the corresponding aggregates of Table~\ref{tab:main-results} up to rounding; serving and all-case rewards use different task subsets.
Prime is the most expensive in both phases (\$0.232 and \$0.142 per task) and reaches the highest test reward on Hidden cases (61.7), while CH reaches 57.7 at 52\% of Prime's test cost (\$0.074).
The two phases can rank a harness differently: SkillOpt and Mem0 spend most of their serving budget on learning, so their test tasks cost less than half of their serving tasks, whereas RAG is the only harness whose test task (\$0.083) costs more than its serving task (\$0.066), because the retrieved exemplars keep lengthening its prompt.
Across models, GLM-5.3 and Kimi K3 buy the highest reward at roughly thirteen to sixteen times the per-task cost of GLM-5.3 Flash, whose test reward on Hidden cases (44.6) trails theirs by 6--8 points.
Table~\ref{tab:cost-correctness} lists every value.

\begin{figure*}[t]
  \centering
  \includegraphics[width=\textwidth]{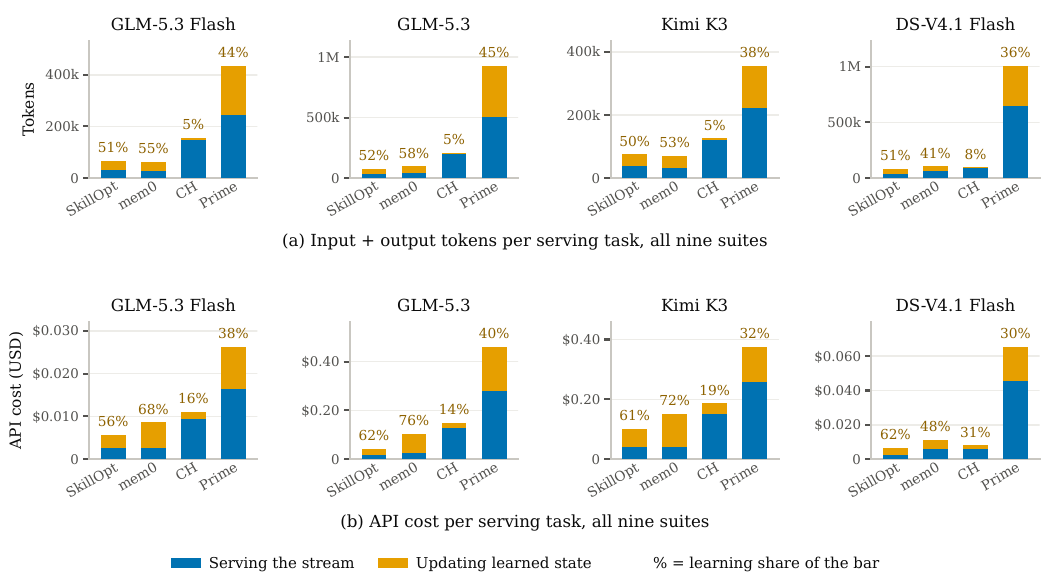}
  \caption{\textbf{Adaptation Cost Analysis.}
  \textbf{(a)} input plus output tokens and \textbf{(b)} API
  cost per serving task, pooled over all nine suites, split between serving
  the stream and updating the learned state. Percentages give the learning
  share of each bar. Prime's learning part is its post-score turn, recovered
  from the session records; its serving part is the rest of the wire ledger's
  serving bucket. Scored test tasks are excluded. Each panel has its own
  scale; RAG (no learning-side compute under BM25) is omitted.}
  \label{fig:run-cost}
  \label{fig:run-cost-combined}
\end{figure*}

The cost of adaptation depends on how a harness divides computation between serving tasks and updating its learned state.
Figure~\ref{fig:run-cost} splits every learning lane's serving-stream spend into the two parts, per serving task, for all four open-weight models and all nine suites.
SkillOpt's learning work is skill optimization and validation, Mem0's is its post-outcome learning turn and memory-extraction model, CH's is its automatic Refiner passes, and Prime's is the post-score turn in which the agent reads the outcome, writes memory and calls \texttt{refine}, measured from its per-ticket session records (Appendix~\ref{app:learning-harnesses}).
The split is a property of the harness more than of the model: learning takes 56--62\% of SkillOpt's API cost and 48--76\% of Mem0's on every model, against 14--31\% for CH and 30--40\% for Prime.
Counting tokens, input (cached or not) plus output, the learning share is 42--58\% for Mem0, 50--52\% for SkillOpt and 36--45\% for Prime, but only 5--8\% for CH, whose serving turns re-send its prompt, memory and skill overviews on every call and are therefore token-heavy but mostly served from cache. Prime consumes the most tokens per task on every model, because each ticket opens a fresh session that re-reads the harness state over 20--29 model calls (24 on average).
Prime is the most expensive harness per task on every model, from \$0.026 on GLM-5.3 Flash to \$0.46 on GLM-5.3; of the four harnesses in Figure~\ref{fig:run-cost}, CH is second on three models and Mem0 on DeepSeek V4.1 Flash.

\FloatBarrier

\subsection{Training-Based Methods}
\label{app:training-based-methods}

We explored supervised fine-tuning (SFT) and GRPO-style reinforcement learning~\citep{shao2024deepseekmath}.
In preliminary trials, 14B and 32B models struggled to infer hidden environment knowledge and complete tasks successfully, even with strong harnesses, making successful self-generated trajectories difficult to obtain.

This limits the correct demonstrations available for success-filtered SFT and the reward contrast available for GRPO under sparse outcome rewards.
When all trajectories in a group receive the same failure reward, their group-relative reward advantages provide no preference among them.
Obtaining informative trajectories is therefore a bottleneck for training from serving experience, before the choice of parameter-update algorithm can become effective.

\subsection{Interpreting Fully Specified Reference Performance}
\label{app:fs-reference-performance}

Fully Specified means that the visible specification is sufficient for the task, not that a model will necessarily trust or follow it.
For example, Opus 5 achieves Fully Specified accuracy of 69.1\%, 52.7\%, and 58.8\% under Blind on Banking L1, L2, and L3, respectively, compared with 100.0\% under Oracle in all three settings (Table~\ref{tab:main-results}).
The Banking prompt warns that actual practice may differ from the documents, and we observe that Opus 5 sometimes speculates about undocumented restrictions without supporting task-specific evidence.
Such speculation can lead it to depart from the visible rules and reduce Blind performance even on Fully Specified tasks.
Oracle's bulletin may provide additional confirmation and discourage these unsupported assumptions.
We therefore interpret changes relative to Blind as differences from the no-learning harness, rather than a pure measure of learning-induced damage.